\documentclass[10pt,journal]{IEEEtran}

\usepackage{stmaryrd}
\usepackage{amsmath}
\usepackage{amssymb}
\usepackage{amsfonts}
\usepackage{graphicx}
\usepackage{url}
\usepackage{float}
\usepackage{mathrsfs,fourier,pifont}
\usepackage{multicol,multienum, multirow}
\usepackage{threeparttable}
\usepackage{algorithmic}
\usepackage{algorithm}
\usepackage{color}
\usepackage{epstopdf}
\usepackage{booktabs}
\usepackage{makecell}
\usepackage{bm}

\usepackage{hyperref}

\usepackage{threeparttable}
\usepackage{booktabs}

\usepackage{relsize}
\let\labelindent\relax
\usepackage{enumitem}

\usepackage{xcolor}

\usepackage{tabularx}

\usepackage{xcolor}
\usepackage{colortbl}
    
\definecolor{first}{RGB}{0,153,0}
\definecolor{second}{RGB}{51,102,255}
\definecolor{third}{RGB}{204,102,0}

\hypersetup{
    colorlinks=true,
    linkcolor=blue,
    filecolor=magenta,
    urlcolor=cyan,
}

\begin{document}
\title{{SAM2Auto: Auto Annotation Using FLASH}}

\author{Arash~Rocky,~\IEEEmembership{Graduate Student Member,~IEEE,}
~Q.~M.~Jonathan Wu,~\IEEEmembership{Senior Member,~IEEE,}

\thanks{Manuscript received ****, ***; accepted *****. Date of publication *****; date of current version *****. Paper no. *****. This work was partially supported by the TrustCAV, a CREATE program of the Natural Sciences and Engineering Research Council of Canada.}

\thanks{A. Rocky and Q.~M.~J.~Wu are with the Department of Electrical and Computer Engineering, University of Windsor N9B 3P4, Canada. (Corresponding author: Q.~M.~J.~Wu. e-mail: \{rocky, jwu\}@uwindsor.ca).}

}


\markboth{Journal of \LaTeX\ Class Files,~Vol.~14, No.~8, Jan~2025}%
{Shell \MakeLowercase{\textit{et al.}}: Bare Demo of IEEEtran.its
for Journals}
\maketitle

\begin{abstract}

Vision-Language Models (VLMs) lag behind Large Language Models due to the scarcity of annotated datasets, as creating paired visual-textual annotations is labor-intensive and expensive. To address this bottleneck, we introduce SAM2Auto, the first fully automated annotation pipeline for video datasets requiring no human intervention or dataset-specific training.
Our approach consists of two key components: SMART-OD, a robust object detection system that combines automatic mask generation with open-world object detection capabilities, and FLASH (Frame-Level Annotation and Segmentation Handler), a multi-object real-time video instance segmentation (VIS) that maintains consistent object identification across video frames even with intermittent detection gaps. 
Unlike existing open-world detection methods that require frame-specific hyperparameter tuning and suffer from numerous false positives, our system employs statistical approaches to minimize detection errors while ensuring consistent object tracking throughout entire video sequences.
Extensive experimental validation demonstrates that SAM2Auto achieves comparable accuracy to manual annotation while dramatically reducing annotation time and eliminating labor costs. The system successfully handles diverse datasets without requiring retraining or extensive parameter adjustments, making it a practical solution for large-scale dataset creation. Our work establishes a new baseline for automated video annotation and provides a pathway for accelerating VLM development by addressing the fundamental dataset bottleneck that has constrained progress in vision-language understanding.
\end{abstract}
\begin{keywords}
Auto Annotation, Object Detection, Multi-Object Tracking, SAM2, Annotation and Segmentation time Handlers, Dataset Creation
\end{keywords}
\section{INTRODUCTION}
\label{sub:Safety}
\IEEEPARstart{I}{MAGE} processing and computer vision community have seen significant progress in recent years \cite{dosovitskiy2021imageworth16x16words}. The emerging of Large Language Models (LLMs) \cite{brown2020languagemodelsfewshotlearners} and Vision Language Models (VLMs) \cite{radford2021learningtransferablevisualmodels} have touched every aspect of humans' lives from assistance in writing an email \cite{touvron2023llamaopenefficientfoundation} to solving scientific problems \cite{bommasani2022opportunitiesrisksfoundationmodels, openai2024gpt4technicalreport, yu2022cocacontrastivecaptionersimagetext, vaswani2023attentionneed}; these models have shown their capabilities and are still on the way of improvement. 

\subsection{Recent Developments in LLMs and VLMs}
\label{sub:CompareLLMVLM}
Having said that, in the comparison of LLMs and VLMs, the latter have not kept the same pace as the former, primarily due to the nature of the datasets they are trained on. LLMs can leverage vast amounts of textual input from diverse and readily available sources on the internet, such as books, articles, and code repositories, enabling large-scale training with minimal manual intervention. This abundance has significantly enriched their performance and scalability \cite{brown2020languagemodelsfewshotlearners}. 
In contrast, VLMs require datasets that pair visual inputs (images and videos) with corresponding textual annotations, such as image captions or object labels. Creating these datasets is labor-intensive, expensive, and time-consuming, leading to a relative scarcity of high-quality, large-scale annotated datasets for VLM training (\cite{radford2021learningtransferablevisualmodels, bommasani2022opportunitiesrisksfoundationmodels}). The annotation process often involves manual human effort or advanced tools, which imposes additional constraints on the scalability of VLMs.
Moreover, the interdisciplinary nature of VLMs, which have to align and bridge two modalities (vision and language), adds complexity to their training compared to LLMs, which operate within the single modality of text. This complexity requires innovative architectures, such as dual encoders or cross-attention mechanisms, to effectively align visual and textual representations (\cite{dosovitskiy2021imageworth16x16words, yu2022cocacontrastivecaptionersimagetext}).
While recent advancements, such as the development of web-scale datasets like LAION \cite{schuhmann2021laion400mopendatasetclipfiltered} and self-supervised learning techniques (e.g., CLIP \cite{radford2021learningtransferablevisualmodels} and DINO \cite{caron2021emergingpropertiesselfsupervisedvision}), have improved the scalability of VLMs, they still lag behind LLMs due to the unmatched richness and diversity of textual data available for training (\cite{radford2021learningtransferablevisualmodels}, OpenAI \cite{openai2024gpt4technicalreport}). However, the progress in vision-language models, as demonstrated by works such as CLIP \cite{caron2021emergingpropertiesselfsupervisedvision}, BLIP \cite{yu2022cocacontrastivecaptionersimagetext}, and Flamingo \cite{bommasani2022opportunitiesrisksfoundationmodels}, shows promise in closing this gap.

\begin{figure}[!t]
\centering
\includegraphics[trim={0.4cm, 0.3cm, 0.5cm, 0.3cm}, clip, width=8.5cm]{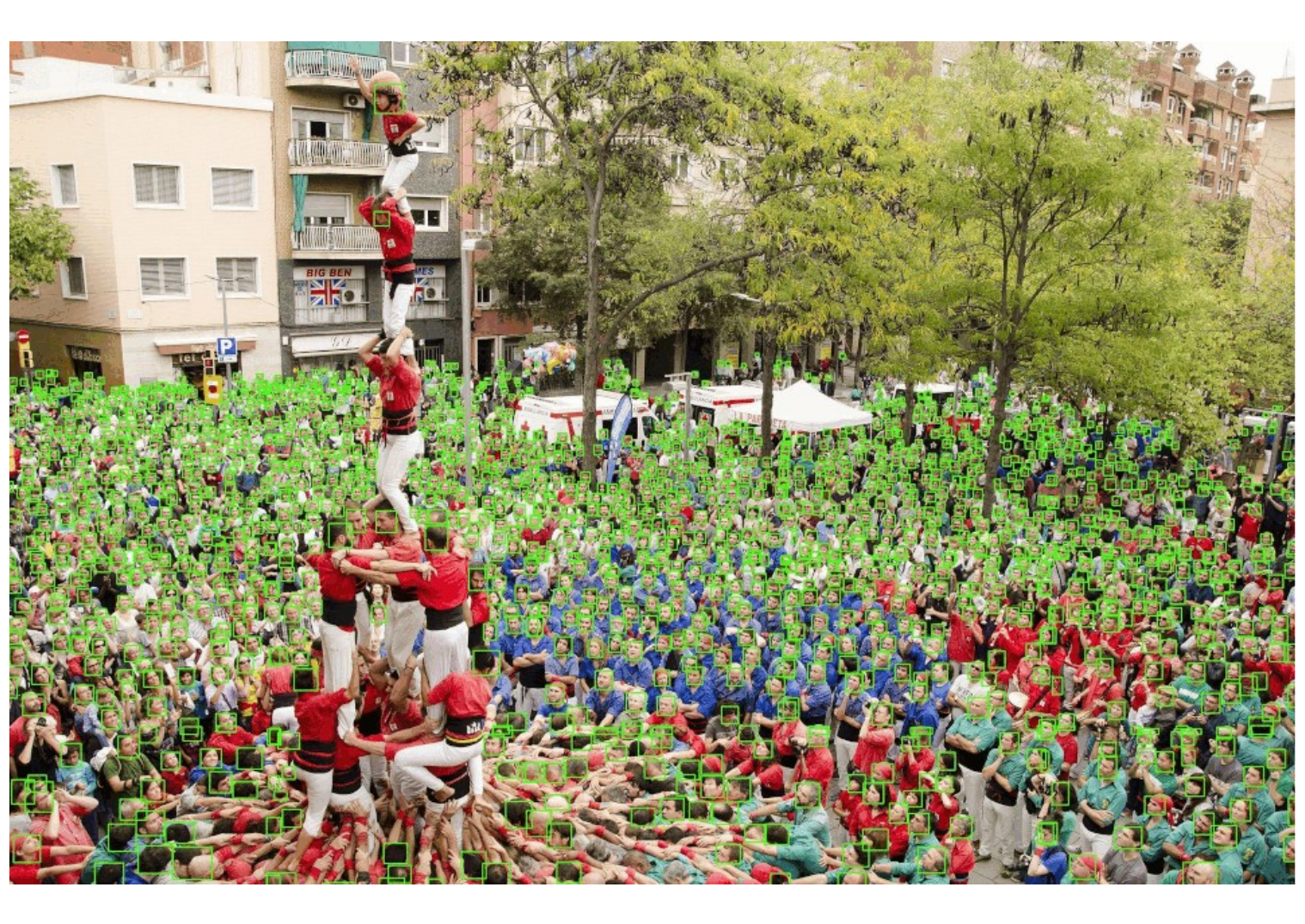}
\caption{An example of detailed annotation in an image\cite{ramirez2024linkedin}.}
\label{fig1}
\vspace{-0.5cm}
\end{figure}

\subsection{Defining the Problem}
\label{sub:Def_Prob}
Vision Transformers (ViTs) have shown remarkable progress in computer vision tasks, yet studies such as \cite{dosovitskiy2021imageworth16x16words} and \cite{touvron2021trainingdataefficientimagetransformers} highlight a critical bottleneck: the requirement for large-scale annotated datasets. This requirement exposes a fundamental challenge in computer vision that remains unresolved: \textbf{Dataset Annotation}. 

The creation of high-quality annotated datasets faces multiple challenges. Manual annotation is a time-consuming process that can extend over months, and is inherently vulnerable to annotator bias and inconsistencies. The financial burden of hiring expert annotators or utilizing data labeling platforms can be substantial, particularly for specialized domains requiring domain expertise. These challenges are further amplified when dealing with complex scenarios, such as annotating small objects or creating precise ground truth annotations for sequential frames, as illustrated in Figure \ref{fig1}. Moreover, as observed in \cite{Arash}, annotations are often tailored to specific methodological requirements, forcing researchers to repeatedly re-annotate datasets to align with their specific research needs---a redundant and resource-intensive process that impedes research progress.

\subsection{Annotation: Term Clarification}
\label{sub:AnnotTerm}
In the context of Machine Learning and Computer Vision, annotation refers to the process of adding structured labels or metadata to raw data (e.g., images, videos, or text) to make it usable for training machine learning models \cite{Everingham2010}. These annotations define the ground truth that the model learns to predict or classify. While text annotation is beyond the scope of this paper, we focus on the relevant aspects of image and video annotation.

Image annotation involves labeling visual data to train models for object recognition and understanding. Common types include bounding boxes for outlining objects, polygonal segmentation for precise shapes, and semantic segmentation to label every pixel \cite{girshick2015fastrcnn}. Additionally, Instance segmentation distinguishes multiple instances of the same class, while keypoint annotation \cite{howard2017mobilenetsefficientconvolutionalneural} marks features like facial landmarks or body joints. Also, 3D annotations label objects in 3D space \cite{Geiger} which are critical for tasks like autonomous driving.

Video annotation extends image annotation to sequential frames, focusing on dynamic elements. Key types include object tracking, which maintains consistent identifications (IDs) across frames, and action recognition, identifying activities like "walking" or "jumping." Event detection labels high-level occurrences (e.g., "traffic accidents" \cite{Arash}), while motion tracking analyzes object trajectories. These annotations are vital for applications like surveillance and autonomous systems \cite{Geiger}.

This research provides a baseline for video annotations, focusing on bounding boxes for objects across sequences with consistent IDs. We argue that automating this type of annotation can pave the way for generating other annotation types with ease.

\subsection{Alternative Approaches}
\label{sub:AltApproaches}
To deal with this deficiency, numerous strategies have been proposed. Researchers of \cite{caron2021emergingpropertiesselfsupervisedvision} and \cite{bommasani2022opportunitiesrisksfoundationmodels} explored self-supervised learning approaches, aiming to reduce dependence on labeled data. Additionally, methods such as online token generation, introduced by \cite{yang2022maskedgenerativedistillation}, and data augmentation combined with regularization techniques, studied by \cite{steiner2022trainvitdataaugmentation}, have been investigated to enhance data efficiency during training.

At the application level, innovative approaches have emerged to further minimize reliance on manual annotations. For example, \cite{dalessandro2024afreecaannotationfreecounting} proposed annotation-free object counting using synthetic data generated by latent diffusion models, coupled with a sorting network trained on ordered image triplets. Similarly, \cite{lu2023openvocabularypointcloudobjectdetection} utilized pre-trained 2D models and contrastive learning to detect and classify 3D objects in point clouds without requiring manual 3D annotations.
Moreover, Emergent Spatial-Temporal Scene Decomposition via Self-Supervision (EmerNeRF) \cite{yang2023emernerfemergentspatialtemporalscene} introduced a self-supervised method to decompose scenes into static and dynamic components which eliminates the need for ground truth annotations or pre-trained models.
As another attempt to avoid 3D annotations, Stereo4D \cite{jin2024stereo4dlearningthings3d} extracts high-quality dynamic 3D motion and long-term motion trajectories from internet stereo videos to create a large-scale, real-world 4D scene dataset.

\subsection{Solution: Auto Annotation}
\label{sub:ActSol}
As discussed in \ref{sub:CompareLLMVLM}, the development of Vision-Language Models (VLMs) has been hindered by the scarcity of annotated datasets. While researchers have proposed creative approaches to address this limitation \ref{sub:AltApproaches}, the field still lacks a robust automatic annotation method that can generate high-quality datasets with minimal human intervention.


We propose that the solution lies in leveraging Open-Vocabulary Object Detection methods 
\cite{zohar2023fomo, YOLO-World, Sapiens, RTDETR, DINO-X, Prompt-guidedDETR, DetCLIP, AligningBagRegions, DetCLIPv2, FMGS, GLEE, Region-Aware, OpenEMMA, OWL, ExploitingUnlabeleddatavision, GroundingDINO}.
These detectors can generate bounding boxes for both known and unknown object classes with high confidence. The key challenge, however, is maintaining consistent object IDs across video frames—a task that requires reliable object tracking with minimal ID switches while achieving a high Intersection over Union (IoU) between detected and ground truth tracks. It is important to note that if auto annotation for sequences of frames (i.e., entire datasets) is achieved, the problem for single-image datasets is effectively resolved as well.

Recent work on SAM 2: Segment Anything in Images and Videos \cite{SAM2} has made progress in this direction by introducing trackability features. Their approach allows human annotators to provide annotations for the first frame and refine subsequent frames using positive and negative clicks. While this reduces manual effort, it still requires human intervention.

Our extensive experiments with open-world object detection models revealed a critical limitation: while these models can theoretically detect target objects without additional training, their performance heavily depends on frame-specific hyper-parameter tuning. Using a single set of parameters across an entire dataset or video sequence leads to numerous false positives or missed detections, particularly in scenes with dynamic backgrounds.

To address these challenges, we introduce SAM2Auto—an Auto Annotation pipeline that combines two key components:

\begin{enumerate}
    \item \textbf{SMART-OD}: A robust object detection system that integrates SAM2 \cite{SAM2} for automatic mask generation with YOLO-World \cite{YOLO-World} for open-world object detection, enhanced by SAHI \cite{SAHI} for improved accuracy. Our system employs statistical approaches to minimize false positives before tracking.
    \item \textbf{FLASH} (\textbf{F}rame-\textbf{L}evel \textbf{A}nnotation and \textbf{S}egmentation \textbf{H}andler): A Multi-Object Tracker (MOT) that can extends trackability features of any memory-based segmentation model to maintain consistent object identification across frames, tracking objects from their first detection until their final appearance regardless of intermittent detection gaps.
\end{enumerate}

This approach delivers several key advantages: it eliminates the need for manual annotation, ensures consistent object identification across frames, and operates without dataset-specific training or extensive parameter tuning. Our experimental results demonstrate that SAM2Auto can automatically generate high-quality annotations in significantly less time than manual approaches while maintaining comparable accuracy.

\noindent
The main contributions of this paper are as follows:
\begin{enumerate}
    \item We introduce SAM2Auto, the first fully automated annotation pipeline for video datasets that requires no human intervention or dataset-specific training
    
    \item We present SMART-OD, a novel object detection system that combines SAM2, YOLO-World, and SAHI to achieve consistent and accurate object detection across varied scenes
    
    \item We develop FLASH, a multi-object real-time video instance segmentation (VIS) that maintains object identity across frames even with intermittent detection gaps
    
    \item We demonstrate through extensive experiments that SAM2Auto achieves comparable accuracy to manual annotation while significantly reducing annotation time and cost
    
    \item We provide comprehensive ablation studies showing the effectiveness of each component and their integration within the pipeline
\end{enumerate}


With the general overview of Auto Annotation that was provided, the rest of this paper is organized as follows: In Section~\ref{S:2}, related works that are close to Auto Annotation subject is discussed. Section~\ref{S:3} entails object detction approach SMART-OD, Object tracker FLASH and overall SAM2Auto architectures. Section~\ref{S:4} provides experimental results and relevant discussion. Future directions are argued in Section~\ref{S:5} Finally, the paper is concluded in Section~\ref{S:6}.

{\section{Related Works}
\label{S:2}}
\subsection{Proper Object Detection for Auto Annotation}
\label{sub:ObjDec}
Several \textbf{Open-Vocabulary Object Detections.} methods have advanced automated annotation, each excelling in specific areas. YOLO-World \cite{YOLO-World} extends the YOLO series with vision-language modeling, enabling efficient zero-shot detection of unseen objects. Its real-time performance (35.4 AP on LVIS at 52 FPS) makes it ideal for large-scale annotation. DINO-X \cite{DINO-X} offers robust open-world detection with superior generalization but at a higher computational cost, limiting real-time usability. Sapiens \cite{Sapiens} specializes in human-centric tasks like pose estimation, excelling in high-resolution human annotations but lacking general applicability. RT-DETR \cite{RTDETR} balances accuracy and speed with multi-scale processing but falls short of YOLO-World’s real-time efficiency. Overall, we integerated YOLO-World into our object detection model as it is the most practical choice for scalable, real-time automated annotation due to its speed, adaptability, and zero-shot capabilities.

\subsection{Handling Illumination Variations in Auto Annotation}
\label{sub:illuVar}
\textbf{Changing in lighting conditions} and diverse weather conditions significantly challenge object detection systems. Traditional approaches address this through data augmentation using datasets like BDD100K \cite{BDD100K} and nuScenes \cite{nuScenes}, or through multi-modal sensor fusion as in MFNet \cite{MultiSensorAdverseWeather}. Low-light enhancement techniques have emerged as crucial preprocessing methods, with solutions ranging from neural network-based approaches like MIRNet \cite{EnrichedFeatures} and RUAS \cite{Retinex}, to zero-reference methods like Zero-DCE \cite{Zero-DCE}. Domain adaptation techniques, such as CycleGAN \cite{CycleGAN}, have also been explored to adapt models trained on daytime images for nighttime conditions. While these techniques improve detection performance, they typically require extensive training data or computationally intensive preprocessing. Our approach leverages SAM2 \cite{SAM2} for automated image segmentation prior to object detection, requiring no additional training while effectively handling illumination and weather variations. This simplifies the pipeline while maintaining robust performance across diverse conditions without the need for domain-specific adaptations.

\subsection{Robust Tracking in Auto Annotation}
\label{sub:RobTrk}
Tracking has evolved significantly in computer vision and image processing. While traditional Multiple Object Tracking (MOT) remains a challenging research area \cite{TrackFormer, BoostTrack, ByteTrack, DiffMOT, GHOSTTracker, ROMOT, Samba, Sparsetrack, SUSHI}, many studies have expanded the scope to Multi-Object Tracking and Segmentation (MOTS) \cite{masa}, Video Object Segmentation (VOS) \cite{JointModeling, PuttingObject, TAM-VT}, Interactive VOS (iVOS) \cite{VideoClick, TrackingAnything, SegmentMeetsPointTracking}, and Video Instance Segmentation (VIS) \cite{OVIS, TowardOVVIS}. 
Tracking has become even more diverse with the emergence of point trackers \cite{DynOMo, SpatialTracker, Track4Gen, CoTracker3}, driven by the need for precise keypoint correspondences in 3D object reconstruction. If we generalize this diversity as \textbf{Tracking Anything in Videos}, we can categorize tracking methods as follows: MOT trackers are based on bounding boxes of objects, trackers of MOTS, VOS, iVOS, and VIS rely on pixel-level masks of objects, while point trackers focus on fine-grained keypoints of objects. 

Despite the lack of large-scale annotated datasets in each category \cite{TrackFormer, Samba, SAM2, CoTracker3}, tracking methods have remained largely independent of dataset in other categories, with little convergence between their methods. As a solution, OmniTracker \cite{OmniTracker} introduced a \textbf{unified tracking-with-detection} architecture to integrate bounding box trackers (MOT) with mask trackers (VOS, iVOS, VIS). However, this method requires dataset-specific training, making it impractical for automated annotation. 

To address this limitation, SPAM \cite{spam2024eccv} was developed for auto-annotation, leveraging synthetic pretraining, pseudo-labeling, and active learning with a graph-based model. While it focuses \textbf{primarily on MOT annotation}, it still requires limited human supervision and additional training. 

\textbf{For VOS and iVOS}, SAM2 \cite{SAM2} introduced a memory bank-based approach for efficient mask tracking. While SAM2 can track objects with well-defined boundaries, including subparts, it struggles with long video sequences and large numbers of objects due to out-of-memory issues. Notably, none of the recent studies on SAM2 \cite{BetterCallSAL, EfficientTrackAnything, ReferEverything, SAMURAI, SMITE} have addressed this limitation.  

To bridge these gaps, we introduce \textbf{SAM2ASH}, a unified bounding box (MOT) and mask-based (VOS, iVOS, VIS) tracker for annotation that extends SAM2’s trackability while eliminating memory limitations. Unlike existing memory bank-based trackers, SAM2ASH efficiently scales to long video sequences and multiple objects. Moreover, it requires \textbf{no additional training}, leveraging the rich SAM2 checkpoints for robust performance. 

\subsection{Review on Auto Annotation}
\label{sub:RevAuto}
One of the earliest works in auto-annotation, 
\cite{MILAutoAnnot}, utilized Multiple Instance Learning (MIL) as a \textbf{weakly supervised method} for image classification and textual annotation alignment. This approach was among the first to recognize the impracticality of full supervision due to large-scale unlabeled datasets. Similarly, 
\textbf{Cost-Efficient Labeling} \cite{LabelingCosts} examined the trade-off between annotation cost and model performance, offering insights into efficient dataset curation. While MIL-based approaches rely on indirect supervision, cost-aware methods focus on balancing performance with limited annotation resources. These methods contrast with fully automated vision-language models, which aim to eliminate manual labeling altogether by leveraging large-scale pretraining.

Recent advancements in vision-language models have enabled \textbf{auto-annotation without requiring predefined categories}. 
\cite{OVODScaling} and 
\cite{DetCLIPv3} utilize vision-language pretraining to detect objects outside traditional labeled datasets. Unlike weakly supervised methods that depend on predefined object categories, these models generalize beyond limited annotations. Expanding on this, Automated Pixel-Level Open-Vocabulary Instance Segmentation (APOVIS) \cite{APOVIS} and 
Real-Time VIS
\cite{RealTimeVOS} extend open-vocabulary capabilities to instance segmentation in both static and video contexts. While vision-language models aim for zero-shot or few-shot annotation, reducing human involvement, their reliance on large-scale pretraining introduces potential biases and inaccuracies, particularly when applied to domain-specific tasks such as autonomous driving.

While vision-language models enable broad generalization, \textbf{autonomous driving applications} require domain-specific annotation techniques. Automatic Data Engine (AIDE) for Object Detection in Autonomous Driving \cite{AIDE} introduces a self-training framework that iteratively refines annotations, reducing errors over time. 
Similarly, \cite{LidarAutoLabel} focuses on LIDAR-based object labeling, refining annotations based on motion trajectories. Unlike open-vocabulary models, which rely on large-scale Internet data, , these methods integrate sensor data for improved domain-specific accuracy. However, this task-specific approach limits their ability to generalize across diverse datasets, a challenge that vision-language models handle more effectively.

Beyond academic research-driven approaches, \textbf{industry tools} such as Cosmos World Foundation Model Platform for Physical AI \cite{CosmosAI} and Auto Labeling Images with Roboflow \cite{RoboflowAutoLabel} integrate foundation models to automate annotation in commercial applications. These platforms provide scalable annotation pipelines but often depend on proprietary data and infrastructure. In contrast, SAM2's data engine \cite{SAM2} has evolved through three key phases to improve video seg-mentation annotation: (1) frame-by-frame annotation using the original SAM, (2) temporal mask propagation through SAM2 Mask, and (3) full integration of SAM2, which incorporates temporal memory and multi-prompt support. This system supports continuous model retrain-ing, quality verification, and automatic masklet generation, ensuring diverse and high-quality annotations. By leveraging a human-in-the-loop approach, SAM2 creates a virtuous cycle that continuously improves efficiency and annotation quality. Unlike traditional industry tools that depend on closed datasets, SAM2’s open integration of temporal consistency and automation makes it particularly suitable for large-scale and high-accuracy video segmentation tasks.

Following industrial solution, recent academic research has explored \textbf{semi-supervised approaches} as an alternative to make annotation even more automated. Semi-Supervised Open-World Object Detection \cite{SemiSupervisedOWOD} and SPAMming Labels \cite{spam2024eccv} leverage semi-supervision to refine labels with minimal human intervention. Unlike weakly supervised and domain-specific annotation methods, which rely on limited super-vision signals, semi-supervised approaches dynamically refine annotations using both labeled and unlabeled data. While they balance annotation cost and model accuracy, they remain less flexible than open-vocabulary models, which can detect unseen objects without retraining.

Despite these advancements, existing methods struggle with either accuracy, efficiency, or adaptability. To address these limitations, we introduce \textbf{SAM2Auto}, a novel framework that combines a fast and precise open-vocabulary object detection system with a pixel-level open-vocabulary VIS framework. \textbf{SMART-OD} serves as a robust open-vocabulary object detector, resilient to illumination variations, while \textbf{FLASH} functions as an MOT real-time VIS capable of recovering inconsistent detections across frames to generate highly accurate instance-level annotations. The name SAM2Auto reflects the dual use of SAM2 in both detection and tracking, reinforcing its role in automated annotation.

\section{Auto Annotation Method}
\label{S:3}
In this section, we introduce SAM2Auto, a novel auto-annotation pipeline that integrates SMART-OD, an open-vocabulary object detection model, and FLASH, a multi-object real-time video instance segmentation (VIS) model, to achieve cost-efficient and highly accurate annotations. By combining these components, SAM2Auto enables high-quality instance-level annotations while maintaining efficiency in large-scale datasets. The following subsections detail each component and its role in the overall system.

\subsection{\textbf{SMART-OD: Dataset-Optimized Pipeline for Robust Open-Vocabulary Detection}}
\label{sub:detection_pipeline}

We propose a multistage object detection and verification pipeline that generalizes open-vocabulary detection capabilities from single images to entire sequences and datasets. This approach, which we call \textbf{S}egmentation, \textbf{M}ask-guided \textbf{A}nalysis and \textbf{R}obust \textbf{T}hresholding for \textbf{O}bject \textbf{D}etection (\textbf{SMART-OD}), addresses a critical limitation in current open-vocabulary methods, which typically excel on individual images but struggle to maintain consistency and accuracy across varied frames in video sequences or diverse images within datasets. As illustrated in Figure \ref{fig:pipeline}, SMART-OD pipeline consists of three sequential stages that progressively refine the detection process, as follows:

\begin{figure}[t]
    \centering
    \includegraphics[trim={0.4cm, 0.3cm, 0.5cm, 0.3cm}, clip, width=8cm]{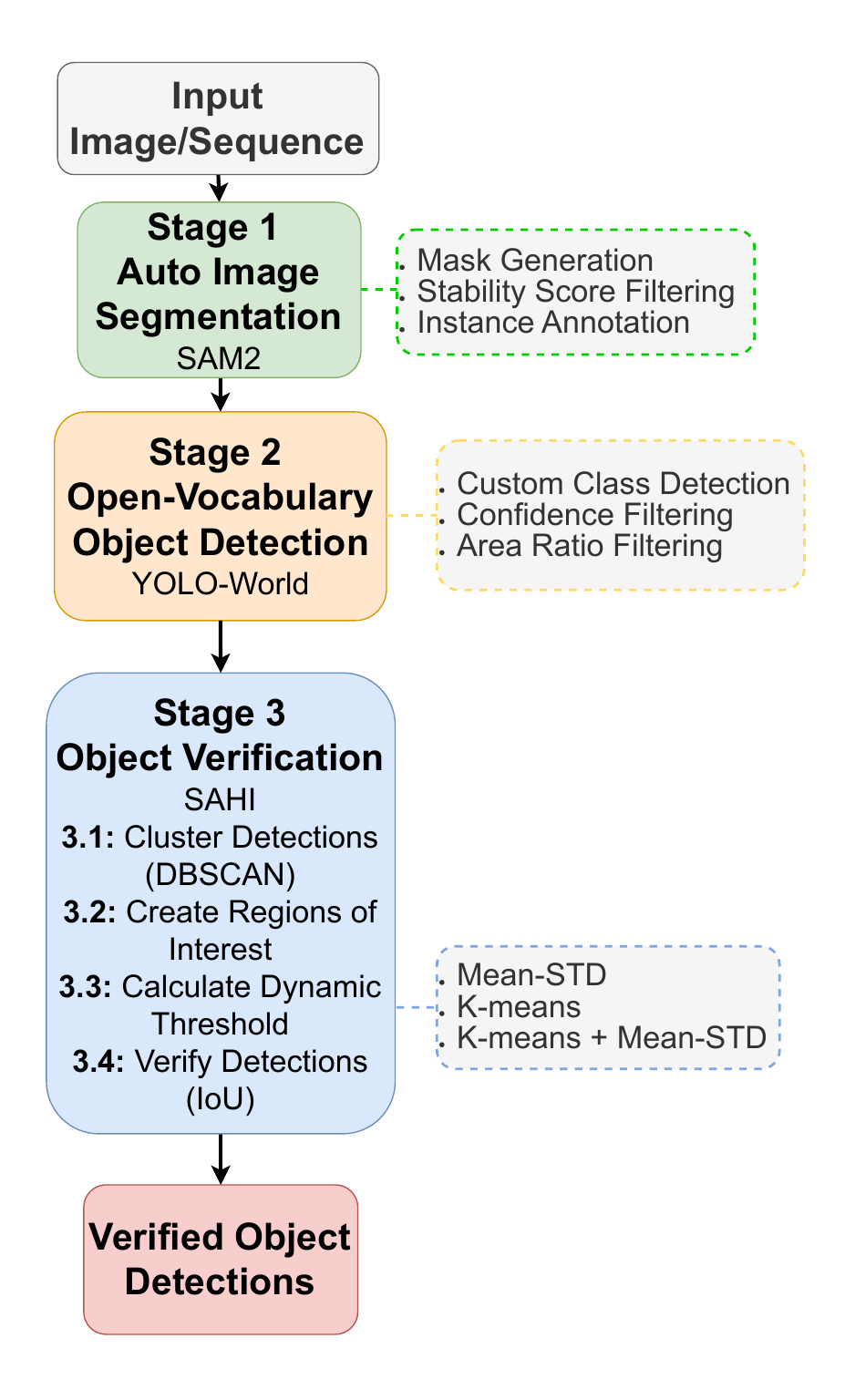}
    \caption{SMART-OD: Dataset-optimized pipeline for robust open-vocabulary detection. The system processes input images through three main stages: Segmentation with SAM2, Mask-guided Analysis with YOLO-World, and Robust Thresholding using SAHI-based verification with dynamic thresholding.}
    \label{fig:pipeline}
\end{figure}

\subsubsection{\textbf{Segmentation through Auto Mask Generation}}
\label{subsub:InsSegSAM2}
The first stage of SMART-OD employs a foundation model for visual segmentation that excels at identifying object boundaries, for which we applied SAM2 \cite{SAM2}. The segmenting model generates instance masks for all potential objects in the scene without semantic classification, which transforms input images with varying illumination conditions into uniform masked representations establishing a consistent visual representation across diverse imaging conditions.




\subsubsection{\textbf{Mask-guided Analysis in Object Detection}}
\label{subsub:YOLO_World}
For the second stage of SMART-OD, we implement an Open-Vocabulary Object Detection (OVOD) approach capable of identifying objects consistently across all sequences and frames of each dataset. As discussed in \ref{sub:ObjDec}, we selected YOLO-World \cite{YOLO-World} as the most suitable OVOD based on its speed, adaptability, and zero-shot capabilities when operating on mask-guided inputs. 






\subsubsection{\textbf{Robust Thresholding in Detection Verification}}
\label{subsub:DetVer}
Although YOLO-World provides efficient open-vocabulary detection, optimizing its parameters to detect the majority of objects across diverse datasets remains challenging. In the SMART-OD pipeline, we address this by implementing a robust verification stage that suppresses inevitable false positives while maintaining high recall. This stage employs SAHI \cite{SAHI} for verification through regional processing and includes several key components:

\vspace{0.2cm}
\noindent \textbf{3.1: Detection Clustering.}
\noindent We cluster spatially proximate detections using DBSCAN \cite{DBSCAN} to create regions of interest (ROIs):

\begin{equation}
    C = \text{DBSCAN}(B, \epsilon, \mu)
    \label{eq4}
\end{equation}

\noindent where $C$ represents the clusters, $B$ is the set of bounding boxes, $\epsilon$ is the maximum distance between points, and $\mu$ is the minimum number of samples in a cluster.
For each cluster, we create a combined ROI:

\begin{equation}
    R_i = [\min_{b \in C_i} x_1, \min_{b \in C_i} y_1, \max_{b \in C_i} x_2, \max_{b \in C_i} y_2]
    \label{eq5}
\end{equation}

\noindent where $R_i$ is the ROI for cluster $C_i$, and $(x_1, y_1, x_2, y_2)$ are the coordinates of each bounding box $b$ in the cluster.

\vspace{0.2cm}
\noindent \textbf{3.2: Dynamic Thresholding.} As the key step in stage 3, the SMART-OD pipeline suppresses false positives by adaptive thresholding that responds to each frame's confidence distribution. The dynamic threshold $\theta_d$ is calculated using one of four methods, then constrained by a minimum threshold $\theta_{\text{min}}$:

\begin{align}
\label{Dynm_thrsh}
\theta_d &=
\begin{cases}
\mu_s - \sigma_s & \text{(Mean-STD)} \\
\min_{s \in S_t} s & \text{(K-means)} \\
\mu_{l} + 2\sigma_{l} & \text{(K-means + Mean-STD)} \\
\text{Two-stage clustering} & \text{(Double K-means)}
\end{cases} \\[0.5em]
\theta_{\text{final}} &= \max(\theta_d, \theta_{\text{min}})
\end{align}

where $\mu_s, \sigma_s$ are the mean and standard deviation of all confidence scores, $S_t$ represents scores in the top two K-means clusters, and $\mu_{l}, \sigma_{l}$ are statistics from the lowest confidence cluster.









\noindent \textbf{3.3: ROI Processing and Verification}

\noindent Each ROI is processed individually using the SAHI approach. For a detection to be verified, it must meet two criteria:

\begin{enumerate}
    \item \textbf{Detection IoU:}
    The detection must have an IoU with a SAHI prediction above the verification threshold:
    \begin{equation}
        \max_{p \in P} \text{IoU}(b, p) > \theta_v
        \label{eq10}
    \end{equation}
    where $P$ is the set of SAHI predictions, $b$ is the detection box, and $\theta_v$ is the verification IoU threshold.

    \item \textbf{Detection Confidence:}
    The detection confidence must exceed the dynamic threshold based on Eq. \ref{Dynm_thrsh}:
    \begin{equation}
        c_b > \theta_{\text{final}}
        \label{eq11}
    \end{equation}
    where $c_b$ is the confidence of detection $b$.
\end{enumerate}

\noindent This dual-criterion approach ensures that only high-confidence detections with strong verification support are retained as the final detections of SMART-OD. The detailed formulas and comprehensive algorithm of SMART-OD for processing every sequence of a dataset is provided in appendix \ref{app:mathematical_formulations}.

\subsection{\textbf{FLASH: Real-Time VIS for Auto-Annotation}}
\label{sub:FLASH}

\begin{figure*}[!t]
\includegraphics[trim={0.5cm, 0.2cm, 0.5cm, 0.1cm}, clip, width=18cm]{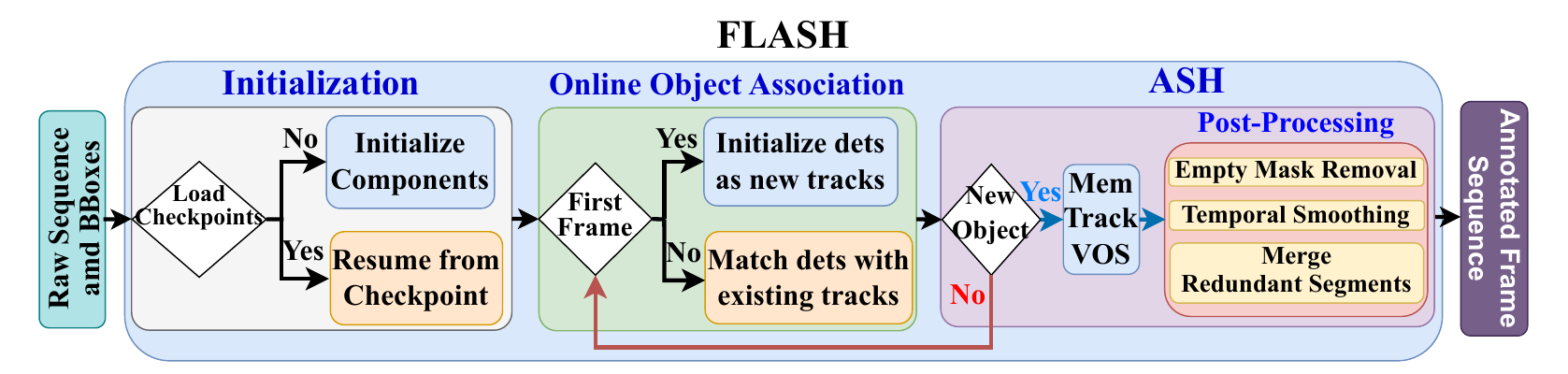}
\caption{FLASH system architecture showing the three main modules: Initialization, Online Object Association, and Annotation and Segmentation Handler (ASH). It gets Raw Sequence of Frames and Bounding Boxes (BBoxes) and outputs Annotated Frame Sequence with the core of Memory-based Segmentation (Mem Track VOS)}
\vspace{-0.3cm}
\label{fig3}
\end{figure*}

We present FLASH (Frame-Level Annotation and Segmentation Handler), a framework integrating multi-object tracking with high-quality segmentation by leveraging a memory-based segmentation module. FLASH transforms bounding box detections into detailed polygon segmentations while maintaining temporal consistency across frames.

FLASH consists of three main modules, as illustrated in Figure~\ref{fig3}:

\begin{enumerate}
    \item \textbf{Initialization Module}: Handles checkpoint loading and system preparation
    \item \textbf{Online Object Association Module}: Associates detections with an online object tracking
    \item \textbf{Annotation and Segmentation Handler (ASH)}: Manages memory-based segmentation model
\end{enumerate}

\subsubsection{\textbf{Initialization Module}}
\label{subsub:initialization}

The Initialization Module serves as the entry point to the FLASH framework, handling system preparation and providing robust resume capabilities for processing long video sequences. This module performs three key functions:

\begin{itemize}
    \item \textbf{Checkpoint management}: Loads previously saved processing states and handles serialization of current state
    \item \textbf{Resume capability}: Enables processing to continue from specific frames after interruptions
    \item \textbf{Memory-efficient processing}: Implements adaptive chunk selection mechanism to handle long videos that exceed memory constraints, utilizing temporal frame optimization to maintain consistency
\end{itemize}

The detailed implementation of checkpoint management protocols and resume capability algorithms are provided in Appendix~\ref{app:Flash_Details}.
To address \textbf{memory limitations} when processing long videos in ASH module, the Initialization module implements an adaptive chunk selection mechanism. This system maintains temporal consistency through optimal frame identification, which selects frames using the following criterion:
\vspace{-0.2cm}
\begin{multline}
\text{OptimalFrame} = \\
\arg\max_{f} |\mathcal{O}_f| \
\text{such that } c-w \leq f \leq c+w
\end{multline}

\noindent Where $\mathcal{O}_f$ represents the set of objects in frame $f$, $c$ is the current frame, and $w$ is the window size.

\subsubsection{\textbf{Online Object Association Module}}
\label{subsub:online_association}

The Online Object Association Module forms the critical bridge between object detection and segmentation in the FLASH framework. This module maintains temporal consistency by systematically associating detection boxes with an online object tracker as our Multi-Object Tracking (MOT) solution, complemented by IoU-based association to preserve consistent object identities.

\noindent The module performs three essential functions:
\begin{itemize}
    \item \textbf{Temporal Context Processing}: Differentiating between first frame and subsequent frame handling
    \item \textbf{Detection-Track Association}: Mapping newly detected objects to existing tracks
    \item \textbf{New Object Identification}: Determining which detections require segmentation initialization
\end{itemize}

\noindent Detailed implementation of temporal context processing and detection-track association algorithms are provided in Appendix~\ref{app:Flash_Details}.

\textbf{New Object Identification:}
After processing all associations, any detection not matched to an existing track is identified as a new object requiring track initialization. Formally, the set of new objects $N_j$ for frame $F_j$ is defined as:

\begin{equation}
N_j = \{d_i \in D_j : i \notin \text{dom}(\mathcal{M})\}
\end{equation}

\noindent where $\text{dom}(\mathcal{M})$ represents the domain of mapping function $\mathcal{M}$.

For each unmatched detection, the system assigns a new unique track ID, initializes a new track, and incorporates this information into the set of known tracks. This systematic approach ensures that all objects are properly tracked throughout the video sequence while maintaining their distinct identities.  These new objects will be passed to ASH module to create their annotation and segmentation for the whole sequence from the current frame.

\subsubsection{\textbf{Annotation and Segmentation Handler (ASH)}}

The core of FLASH is the Annotation and Segmentation Handler (ASH) that is initialized with bounding boxes of new objects from the Online Object Association module and employs a memory-based segmentation model to propagate mask segmentations of all these new objects throughout the sequence. This creates a multi-object real-time video instance segmentation (VIS) model. The architecture effectively handles the complex challenges of maintaining temporal consistency, managing computational resources, and balancing accuracy with real-time performance requirements.

The ASH module processes video sequences frame-by-frame while maintaining object identity across temporal boundaries, generating both polygon-based segmentation masks and bounding box representations.

The module performs three essential functions:
\begin{itemize}
    \item \textbf{Memory-Efficient Processing}: Optimizing frame and object handling for computational efficiency (Details can be found in Appendix \ref{app:Flash_Details})
    \item \textbf{Transforming VOS to Real-Time VIS}: Converting single-object segmentation to multi-object tracking
    \item \textbf{Post-Processing}: Enhancing segmentation quality and consistency
\end{itemize}

\textbf{Transforming VOS to Real-Time VIS:}
ASH extends single-object video segmentation capabilities to multi-object tracking through two key mechanisms:

\textit{Integration with Online Object Association:} A segmentation model requires a unique object id for bounding box of new objects. For this requirement, ASH takes advantage of tracking ids that were assigned to relevant objects from the Online Object Association Module. 
More specifically, the set of new objects $\mathcal{N}_t = \{o_1^t, o_2^t, ..., o_n^t\}$ also represents the set of tracked objects from the Online Object Association module, where each object $o_i^t$ is associated with a unique tracking ID. As a result, for a new object detected at frame $t$:

\begin{equation}
o_i^t = \begin{cases} 
\text{track}_j & \text{if}\ \exists j : \text{IoU}(b_i^t, b_j^{t-1}) > \tau_\text{track} \\ 
\text{ID}_\text{new} & \text{otherwise} 
\end{cases}
\end{equation}

\noindent where $b_i^t$ is the bounding box of object $i$ at frame $t$, $\tau_\text{track}$ is the tracking threshold, and $\text{ID}_\text{new} = \max(\text{IDs}) + 1$ for new object assignment. Such an approach maintains consistent object identities across frames, which is a critical component for tracking accuracy.

\textit{Multi-Object Mask Propagation and Representation Conversion:} New objects are fed to the segmentation model which produces binary masks of objects throughout the sequence. For each object $o_i^t$ initialized at frame $t$, the segmentation model generates masks across subsequent frames:

\begin{equation}
    M_i^{t+\delta} = \phi(o_i^t, F_t, F_{t+\delta}), \quad \delta \in \{0, 1, ..., (T-t)\}
\end{equation}

\noindent where $\phi$ represents the memory-based segmentation function that propagates the initial object to subsequent frames.

Our module employs a dual representation strategy for segmentation results by converting binary masks to polygon representations:

\begin{equation}
    P_i^j = \{(x_1, y_1), (x_2, y_2), ..., (x_m, y_m)\} = \psi(M_i^j)
\end{equation}

\noindent where $\psi$ is the contour extraction function that identifies the boundary points of the mask. This conversion offers significant storage efficiency and facilitates downstream processing tasks such as visualization and analysis.

\textbf{Post-Processing:}
ASH applies three crucial post-processing techniques to enhance segmentation quality:

\textit{Empty Mask Removal:} We first identify the temporal terminus of each object $\tau(o_i)$ as the last frame where the object maintains a valid mask representation:
$\tau(o_i) = \max\{t \in [1,T] : \sum_{x,y} M_i^t(x,y) > \epsilon\}$
where $\epsilon$ is a minimal threshold for valid mask content. Subsequently, we perform mask pruning across the sequence by applying a temporal consistency filter $\phi$ defined as:
\begin{equation}
    \phi(M_i^t) = \begin{cases}
    M_i^t & \text{if}\ t \leq \tau(o_i) \\
    \emptyset & \text{if}\ t > \tau(o_i)
    \end{cases}
\end{equation}

\noindent This effectively eliminates phantom masks that occur after an object has exited the scene. The refined set of segmentations $\hat{\mathcal{M}}$ is then computed as:
$\hat{\mathcal{M}} = \{\phi(M_i^t) : \forall i \in [1,n], t \in [1,T] \text{ where } \phi(M_i^t) \neq \emptyset\}$

\textit{Temporal Smoothing:} When segmentation models process each frame separately, they can produce jittery, flickering boundaries due to subtle changes in lighting, viewpoint, or object deformation, even when the actual object moves smoothly. This visual instability not only degrades the perceived quality of the segmentation but can also negatively impact downstream applications like object tracking. By applying temporal smoothing, represented mathematically as a weighted average of current and previous frame segmentations (\ref{eq:smooth}),
the system suppresses high-frequency noise while preserving legitimate motion. This smoothing improves tracking accuracy by reducing false associations caused by erratic boundary changes, and enhances the overall robustness of the system against temporary segmentation failures in individual frames.

\begin{equation}
\label{eq:smooth}
    \hat{P}_i^t = \alpha \cdot P_i^t + (1-\alpha) \cdot P_i^{t-1}
\end{equation}

\noindent where $\alpha \in [0,1]$ is the temporal smoothing factor. 

\begin{figure*}[!t]
\centering
\includegraphics[trim={0.5cm, 0.2cm, 0.5cm, 0.1cm}, clip, width=18cm]{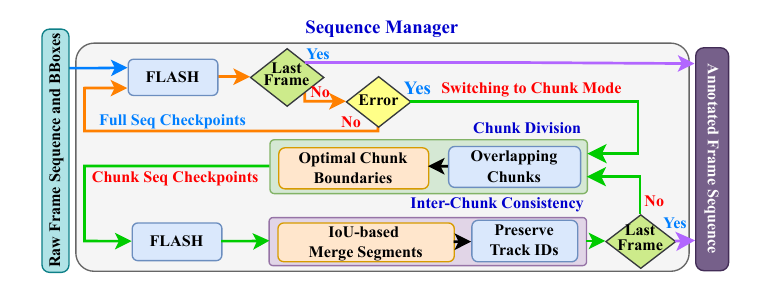}
\caption{FLASH Integration in Sequence Manager: By leveraging Chunk Division, Inter-Chunk Consistency, and Chunk Sequence Checkpoints, FLASH can efficiently handle sequences of arbitrary length and object count without constraints.}
\vspace{-0.3cm}
\label{fig4}
\end{figure*}

\textit{Merging of Redundant Segments:} When segmentation masks are propagated across video frames, they often accumulate errors due to factors such as object deformation, occlusion, lighting variations, model limitations, and the generalized configuration of the Online Object Association module across diverse datasets and sequences. These challenges can result in multiple segmentation instances representing the same physical object—particularly during object splits, merges, or temporary tracking failures. Such redundancy increases computational and storage costs, and may introduce visual artifacts in the final output.
To address this, ASH performs merging of redundant segments, combining those with high overlap based on an Intersection-over-Union (IoU) threshold $\tau_\text{merge}$. This significantly improves both the visual quality and accuracy of object representations. The approach is conceptually similar to non-maximum suppression in object detection but extended to the spatiotemporal domain, ensuring each object is represented by a single, high-quality segmentation mask throughout the video. Thus, the refined set of polygons is computed as:
\begin{equation}
    \hat{\mathcal{P}} = \left\{\hat{P}_i^t : \forall i,t \text{ such that } \max_{j \neq i}\left(\text{IoU}(\hat{P}_i^t, \hat{P}_j^t)\right) < \tau_\text{merge} \right\}
\end{equation}

\noindent where $\tau_\text{merge}$ defines the overlap threshold for merging. Combined with temporal consistency enforcement and noise reduction, this post-processing step significantly enhances segmentation quality and benefits downstream tasks such as object tracking.




\subsection{\textbf{Robust Chunk-Based Processing for Long Sequences}}

FLASH implements an adaptive chunking strategy to handle long video sequences that exceed memory constraints or require robust fallback mechanisms. This approach divides video processing into manageable segments while maintaining temporal consistency across chunk boundaries.

The chunk-based processing subsystem consists of two primary components:

\begin{itemize}
    \item \textbf{Chunk Manager}: Coordinates the division of long sequences into overlapping chunks
    \item \textbf{Chunk Processor}: Handles individual chunk processing with boundary consistency
\end{itemize}

\textbf{Chunk Manager:} 
The system first attempts to process a complete sequence using unified processing. If this fails due to memory constraints or computational limitations, it automatically switches to chunk-based processing. The chunk manager employs a sliding window approach with configurable chunk size and overlap parameters:

\begin{equation}
\mathcal{C} = \{C_1, C_2, ..., C_n\} \text{ where } C_i = [s_i, e_i]
\end{equation}

\noindent where $\mathcal{C}$ represents the set of chunks, $C_i$ represents the $i$-th chunk spanning frames $s_i$ to $e_i$, and chunks maintain temporal overlap to ensure consistency:

\begin{equation}
s_{i+1} = e_i - \omega \text{ for } i \in \{1, 2, ..., n-1\}
\end{equation}

\noindent with $\omega$ representing the overlap size. For optimal chunk boundaries, the system identifies regions with high object density using:

\begin{equation}
\text{OptimalStart}_i = \arg\max_{f \in [s_i-\delta, s_i+\delta]} |\mathcal{O}_f|
\end{equation}

\noindent where $\mathcal{O}_f$ is the set of objects in frame $f$ and $\delta$ defines the search window size.

\textbf{Chunk Processor:} 
For each chunk $C_i$, the processor handles detection-track association and segmentation with special consideration for overlap regions. The Annotation and Segmentation Handler (ASH) is initialized with the object state from previous chunks in the overlap region, ensuring continuity:

\begin{equation}
\mathcal{S}_{i+1}(s_{i+1}) = \mathcal{S}_{i}(e_i - \omega)
\end{equation}

\noindent where $\mathcal{S}_{i}(f)$ represents the segmentation state at frame $f$ in chunk $i$.

\textbf{Inter-Chunk Consistency:} 
For overlap regions between chunks, FLASH applies a specialized polygon-based object merging strategy that resolves identity conflicts through direct IoU matching of the segmentation masks themselves, rather than relying on the looser constraints of the Online Object Association module. This approach, implemented in the \texttt{merge\_overlapping\_segments} function, calculates precise IoU between object polygons across frames in the overlap region:

\begin{equation}
\text{IoU}(O_i^A, O_j^B) = \frac{1}{|F_\text{overlap}|}\sum_{f \in F_\text{overlap}} \frac{|M(O_i^A, f) \cap M(O_j^B, f)|}{|M(O_i^A, f) \cup M(O_j^B, f)|}
\end{equation}

\noindent where $O_i^A$ represents object $i$ from chunk $A$, $O_j^B$ represents object $j$ from chunk $B$, $F_\text{overlap}$ is the set of overlapping frames, and $M(O, f)$ is the mask for object $O$ at frame $f$.

The system maintains a mapping between chunk-specific object IDs and consistent global IDs based on maximum average IoU across the overlap region:

This robust chunking strategy enables FLASH to process arbitrarily long video sequences with bounded memory requirements while maintaining temporal consistency across the entire sequence, including chunk boundaries. Importantly, it eliminates the need for dataset-specific tuning of

\begin{equation}
\text{GlobalID}(O_i^B) = \begin{cases} 
\text{ID}(O_j^A) & \text{if}\ \exists j : \text{IoU}(O_i^B, O_j^A) > \tau_\text{overlap} \\ 
\text{ID}(O_i^B) & \text{otherwise}
\end{cases}
\end{equation}



\noindent tracking parameters, making the approach more robust and generalizable for auto-annotation across diverse video datasets. Furthermore, FLASH incorporates a fallback mechanism that automatically switches between full-sequence processing and chunk-based processing depending on the available computational resources and the complexity of the video sequence. The overall integrated FLASH in Sequence Manager architecture is drawn in Fig. \ref{fig4}.



\subsection{\textbf{SAM2Auto: Unified Pipeline for Auto-Annotation}}
\label{sub:SAM2Auto}
Taking advantage of the SMART-OD object detection framework with the FLASH video instance segmentation system, we introduce a novel auto-annotation pipeline, called SAM2Auto. This pipeline addresses the challenge of creating high-quality instance-level annotations for large-scale datasets with minimal manual intervention. Through a systematic approach to parameter optimization and validation, SAM2Auto achieves robust performance across diverse visual conditions while maintaining computational efficiency. We detail the implementation methodology and provide a structured workflow for deploying this system on arbitrary datasets.
Our approach combines SAM2 \cite{SAM2}, an enhanced version of the Segment Anything Model, for Auto Image Segmentation stage of SMART-OD with YOLO-World \cite{YOLO-World} for open-vocabulary detection stage, and SAHI \cite{SAHI} for verification stage. As for Online Object Association Module of FLASH, we apply ByteTrack \cite{ByteTrack} and for a memory-based video instance segmentation framework of FLASH we apply SAM2 \cite{SAM2}. This integration enables cost-efficient and highly accurate annotations at scale.

\begin{algorithm}
\caption{SAM2Auto: Systematic Auto-Annotation Pipeline}
\label{alg:sam2auto}
\begin{algorithmic}[1]
\REQUIRE Dataset $\mathcal{D}$, Parameters $\Theta$
\ENSURE Annotated Dataset with instance segmentation
\STATE $S_{rep} \gets \arg\max_{S \in \mathcal{D}} \max_{f \in S} |O_f|$ \hfill $\triangleright$ Select sequence with highest object density
\STATE $f_{crowd} \gets \arg\max_{f \in S_{rep}} |O_f|$ \hfill $\triangleright$ Find most crowded frame
\STATE $\Theta_{opt} \gets \arg\max_{\Theta} J(\Theta, f_{crowd})$ \hfill $\triangleright$ Optimize parameters
\STATE Apply SMART-OD with $\Theta_{opt}$ to $S_{rep}$ and evaluate Precision and Recall
\STATE Select random sequence $S_{val}$ for cross-validation
\STATE Verify $\min(P_{val}, R_{val}) \geq \gamma \cdot \min(P_{rep}, R_{rep})$
\FOR{each sequence $S_i$ in dataset $\mathcal{D}$}
    \STATE $D_i \gets \text{SMART-OD}(S_i, \Theta_{opt})$ \hfill $\triangleright$ Apply object detection
    \STATE $V_i \gets \text{SequenceManager}(S_i, D_i, \Theta_{SM})$ \hfill $\triangleright$ Apply Sequence Manager with FLASH
\ENDFOR
\STATE Perform Quality Assurance on stratified sample of sequences
\STATE $Q_i \gets \text{IoU}(V_i, M_i)$ for sampled $S_i$
\FOR{each $S_i$ where $Q_i < \tau_{QA}$}
    \STATE Refine parameters and reprocess $S_i$
\ENDFOR
\RETURN Annotated Dataset with instance segmentation
\end{algorithmic}
\end{algorithm}

\textbf{Integration Methodology:}
SAM2Auto employs a systematic seven-step deployment approach formalized in Algorithm~\ref{alg:sam2auto}: (1) representative sequence selection using maximum object density, (2) parameter optimization with SMART-OD on the most crowded frame, (3) sequence-level verification, (4) cross-sequence validation, (5) dataset-wide detection, (6) full-scale Sequence Manager application with FLASH, and (7) quality assurance through stratified sampling. 

The algorithm implements step (1) by identifying the sequence containing the frame with highest object count, followed by step (2) parameter optimization on this most challenging frame. Step (3) applies the optimized configuration to the entire representative sequence for verification, while step (4) performs cross-sequence validation to ensure generalization across different parts of the dataset. Steps (5) and (6) are handled in the main processing loop, where each sequence undergoes dataset-wide detection followed by Sequence Manager application with integrated FLASH segmentation. Finally, step (7) implements quality assurance through IoU-based evaluation and targeted refinement for sequences below quality thresholds.
The detailed seven-step implementation procedure is provided in Appendix~\ref{appendix:sam2auto_details}.

\section{{EXPERIMENTS}}
\label{S:4}

\subsection{Datasets and Metrics}
\label{sub:DataMet}
\noindent \textbf{Datasets.} To ensure that our method is comparable with recent work in tracking and annotation, we apply it to four renowned object tracking datasets as follows:

\begin{itemize}[leftmargin=*]
    \item \textbf{MOT17} \cite{MOT17} is a standard tracking benchmark with 14 diverse sequences featuring varying camera motions, viewpoints, and crowd densities. Following \cite{GHOSTTracker, SimpleReID, ArTIST, CenterTrack},  for MOT17 public detections, we took advantage of CenterTrack's refined bounding boxes.
    \item \textbf{MOT20} \cite{MOT20} is a challenging benchmark of 8 high-density video sequences captured in real-world crowded scenes. It emphasizes robust tracking under extreme occlusions, tight pedestrian spacing, and limited target visibility, making it ideal for evaluating performance in heavily congested environments.
    \item \textbf{DanceTrack} \cite{DanceTrack} is a multi-object tracking benchmark containing 100+ annotated dance videos where performers display complex movements with frequent occlusions and synchronization. It tests trackers' ability to maintain identity through dynamic motion, unlike standard pedestrian datasets.
    \item \textbf{BDD100K} \cite{BDD100K} is a large-scale driving video dataset with 100,000 annotated clips captured across diverse locations, weather, and lighting conditions. It supports multiple tasks including object detection, segmentation, lane detection, and multi-object tracking. With rich annotations and broad scenario coverage, BDD100K is widely used for training and evaluating models in autonomous driving and multi-task learning. Due to dataset publicity constraints, we used the validation set for evaluation across all methods to ensure a fair comparison.
\end{itemize}

\noindent \textbf{Evaluation Metrics.} In order to provide complementary perspectives on tracking quality, ensuring both accurate detection and consistent identity preservation - crucial for reliable automated annotation systems, we assess our method through following metrics:
\begin{itemize}[leftmargin=*]
    \item \textbf{MOTA} \cite{MOTA}: Measures overall tracking performance by counting errors (false positives, missed detections, ID switches) relative to ground truth objects. While comprehensive, it emphasizes detection quality but undervalues identity preservation.
    \item \textbf{IDF1} \cite{IDF1}: Focuses on identity preservation by measuring how consistently object identities are maintained throughout tracking sequences. Unlike MOTA, it prioritizes consistent identity tracking but might undervalue detection accuracy, as maintaining fewer but consistent tracks can yield higher scores.
    \item \textbf{HOTA} \cite{HOTA}: Balanced metric that equally weights detection accuracy and association quality, providing intuitive evaluation of both components. It addresses limitations of older metrics by offering a more balanced assessment suitable for long-term tracking evaluation.
\end{itemize}

\subsection{Implementation Details}
\label{sub:ImpDet}
\noindent \textbf{Public and Private Detections.} For assessing the tracking capability of FLASH, we applied it to both public and private sets of MOT17, MOT20, and to private set of DanceTrack, as well as the validation set of BDD100K. In case of public sets of MOT17, we took advantage of refined detections of CenterTrack \cite{SimpleReID, ArTIST, CenterTrack, GHOSTTracker}, while for public sets of MOT20, the Tracktor refined detections were utilized \cite{tracktor, MPNs, ArTIST, GHOSTTracker}. On the other hand, for the private sets of MOT17 and MOT20, test sets of DanceTrack and validation sets of BDD100K, we applied YOLO-X detections \cite{YOLOX} considering ByteTracker's training procedure \cite{ByteTrack}.


\noindent \textbf{Architecture.} Our SMART-OD pipeline integrates three state-of-the-art frameworks: Segment Anything Model 2 (SAM2) \cite{SAM2} for high-quality instance segmentation, YOLO-World \cite{YOLO-World} for efficient open-vocabulary detection, and Slicing Aided Hyper Inference (SAHI) \cite{SAHI} for robust verification. This multi-stage architecture transforms input images with varying illumination conditions into uniform masked representations, enabling consistent object detection across entire datasets.
As for our FLASH framework, the Online Object Association Module is leveraging ByteTracker \cite{ByteTrack} with optimized parameters of track thresh is 0.6, match thresh is 0.7, and the Annotation and Segmentation Handler (ASH) employs SAM2 \cite{SAM2} as its memory-based segmentation backbone. We refer to this specific configuration combining SAM2 with ASH as SAM2ASH. This zero-shot system requires no training on target datasets and maintains object identity through sophisticated IoU-based matching.
For memory efficiency, FLASH implements subset frame processing and batch-based object propagation (5-10 objects per batch), enabling processing of arbitrary-length sequences with bounded memory usage. The system automatically shifts between full-sequence and chunk-based processing (50-frame chunks with 10-frame overlap) based on resource availability and sequence complexity.

\begin{figure*}[!t]
\includegraphics[trim={0.2cm, 0.1cm, 0.2cm, 0.1cm}, clip, width=18cm]{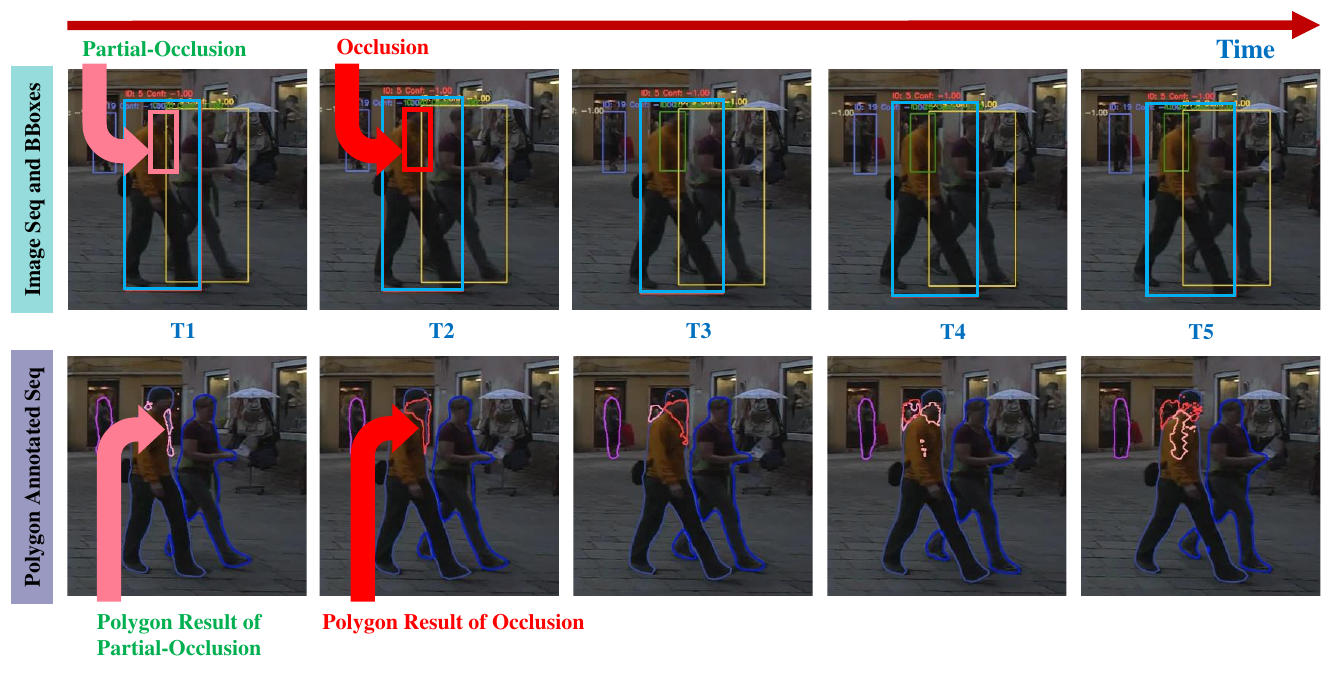}
\caption{Without Online Association module, when bounding boxes from standard MOT at occlusion times are directly fed to FLASH, artifact segments are inevitably generated, leading to degraded overall performance.}
\vspace{-0.3cm}
\label{fig6}
\end{figure*}

\subsection{FLASH as a Multi Object Tracker}

For detailed implementation specifics including parameter configurations, and experimental setup, please refer to Appendices \ref{appendix:experiments_details}.

\subsubsection{\textbf{Ablation Study: Component-wise Analysis of FLASH Framework Performance}}

The FLASH framework integrates several key modules that work in concert to deliver high-quality video instance segmentation. To understand the contribution of each component, we conducted an ablation study analyzing how the removal or modification of specific modules affects the overall system performance.

\noindent \textbf{Importance of the Initialization Module}
The Initialization Module is critical for robust processing of lengthy video sequences. Without this module operating in the processing loop, system failures (such as out-of-memory errors due to high parameter settings of memory-based segments) would require restarting the entire process from the beginning. The checkpoint management system implemented by this module provides essential recovery capabilities by:

\begin{enumerate}
    \item Maintaining serialized system states at strategic points during processing
    \item Enabling resume capabilities from specific frames after interruptions
    \item Implementing the three-phase checkpoint protocol that ensures data integrity
\end{enumerate}

\noindent The robust checkpoint mechanism provides significant value when processing long video sequences or tracking numerous objects that exceed memory constraints or require interrupted workflows. By enabling continuation from the last saved state rather than forcing a complete restart, it dramatically improves efficiency for large-scale video annotation tasks.
It is worth noting that if the method switches from Full Sequence processing to Chunk mode, the system needs to start from the beginning of the sequence as Full Sequence checkpoints do not adapt with chunk mode ones.

\noindent \textbf{Critical Role of the Online Object Association Module}
Removing the Online Object Association Module would introduce two substantial problems:

\begin{enumerate}[leftmargin=*]
    \item \textbf{Computational Inefficiency}: Without this module filtering redundant objects between consecutive frames, the system would process many duplicate objects, dramatically increasing computational costs. These redundancies would generate numerous superfluous polygon segments that would need removal during post-processing, adding further computational overhead.
    
    \item \textbf{Artifact Generation in Occlusion Scenarios}: As clearly illustrated in Figure~\ref{fig6}, when objects become partially or fully occluded (shown at times T1 and T2), standard MOT trackers still generate bounding boxes even for obscured objects. Without the object association module to maintain temporal consistency, the memory-based segmentation would create inconsistent artifact polygons that do not properly align with actual objects. This results in different segmentation patterns for the same object across consecutive frames, severely degrading overall performance.
\end{enumerate}

\begin{figure*}[!t]
\includegraphics[trim={0.2cm, 0.1cm, 0.2cm, 0.1cm}, clip, width=18cm]{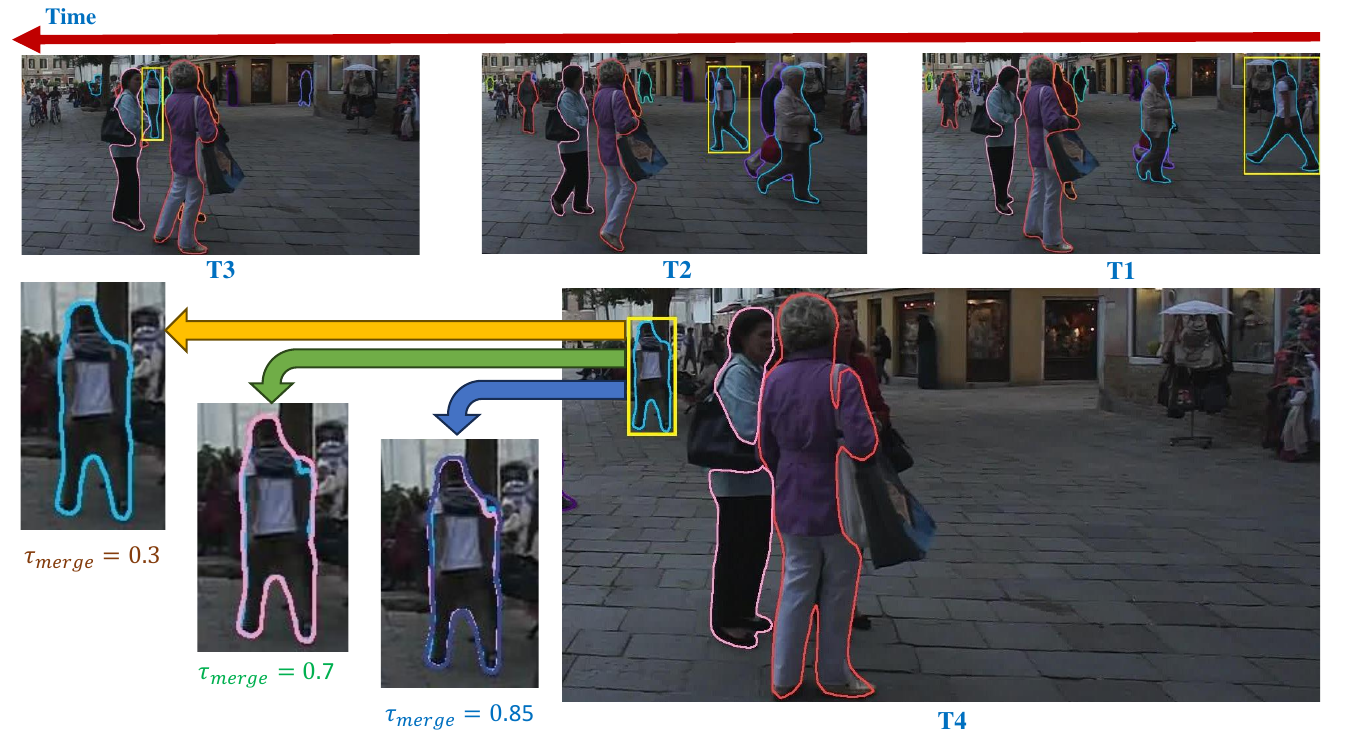}
\caption{Key detection metrics for YOLO-World, SAM2-YW, and SMART-OD on MOT17 training set. SMART-OD achieves the highest precision (0.728) and MOTA (-0.014) by optimizing the precision-recall trade-off.}
\vspace{-0.3cm}
\label{fig7}
\end{figure*}

\noindent \textbf{Necessity of ASH's Post-Processing Steps}
The ASH module's post-processing capabilities, particularly the merging of redundant segments, provide essential refinement of segmentation results. As demonstrated in Figure~\ref{fig7}, varying the $\tau_\text{merge}$ parameter has significant impact:

\begin{enumerate}
    \item With $\tau_\text{merge} = 0.3$, optimal segment merging is achieved
    \item With $\tau_\text{merge} = 0.7$, some redundancies are eliminated
    \item With $\tau_\text{merge} = 0.85$, many redundant segments persist
\end{enumerate}

\noindent This highlights that even though the online association parameters are intentionally set loose to minimize computational overhead, the post-processing steps effectively handle remaining redundancies. The temporal smoothing and merging operations ensure consistent object representations throughout the video sequence, addressing the inevitable imperfections that arise during frame-by-frame processing.

These ablation experiments collectively demonstrate how each component of the FLASH architecture contributes to creating a robust, memory-efficient system for video instance segmentation that maintains temporal consistency even in challenging scenarios with occlusions and limited computational resources.

\begin{table*}[htbp]
\centering
\renewcommand{\arraystretch}{1.2}  
\caption{MOT Performance comparison on MOT17 Public and MOT20 Public detections}
\begin{tabularx}{\textwidth}{lXXXXXXXX}
\toprule
\multirow{2}{*}{Method} & \multicolumn{4}{c}{MOT17 Public} & \multicolumn{4}{c}{MOT20 Public} \\
\cmidrule(lr){2-5} \cmidrule(lr){6-9}
& HOTA $\uparrow$ & MOTA $\uparrow$ & IDF1 $\uparrow$ & IDSW $\downarrow$ & HOTA $\uparrow$ & MOTA $\uparrow$ & IDF1 $\uparrow$ & IDSW $\downarrow$ \\
\midrule
\textbf{FLASH}$^{\ddagger}$ & 43.6 & 34.3 & 57.0 & \textcolor{third}{\textbf{1283}} & 39.8 & 24.2 & \textcolor{third}{\textbf{52.0}} & 2196 \\
GHOST$^{\ddagger}$ \cite{GHOSTTracker} & \textcolor{first}{\textbf{50.7}} & \textcolor{second}{\textbf{61.6}} & \textcolor{first}{\textbf{63.5}} & 1715 & - & - & - & - \\
GHOST$^{\dagger}$ \cite{GHOSTTracker} & 47.4 & 56.5 & \textcolor{second}{\textbf{60.6}} & \textcolor{first}{\textbf{1144}} & \textcolor{first}{\textbf{43.4}} & \textcolor{second}{\textbf{52.7}} & \textcolor{first}{\textbf{55.3}} & \textcolor{first}{\textbf{1437}} \\
ArTIST-C$^{\ddagger}$ \cite{ArTIST} & \textcolor{second}{\textbf{48.9}} & \textcolor{first}{\textbf{62.3}} & 59.7 & 2062 & - & - & - & - \\
ArTIST$^{\dagger}$ \cite{ArTIST} & - & - & - & - & \textcolor{third}{\textbf{41.6}} & \textcolor{first}{\textbf{53.6}} & 51.0 & \textcolor{second}{\textbf{1531}} \\
CenterTrack$^{\ddagger}$ \cite{CenterTrack} & \textcolor{third}{\textbf{48.2}} & \textcolor{third}{\textbf{61.5}} & 59.6 & 3039 & - & - & - & - \\
Tracktor v2$^{\dagger}$ \cite{tracktor} & 44.8 & 56.3 & 55.1 & 1987 & \textcolor{second}{\textbf{42.1}} & \textcolor{third}{\textbf{52.6}} & \textcolor{second}{\textbf{52.7}} & \textcolor{third}{\textbf{1648}} \\
TrackPool$^{\dagger}$ \cite{TrackPool} & - & 55.9 & \textcolor{third}{\textbf{60.5}} & \textcolor{second}{\textbf{1188}} & - & - & - & - \\
UNS$^{\dagger}$ \cite{UNS} & 46.4 & 56.8 & 58.3 & 1914 & - & - & - & -\\
SORT \cite{SORT} & - & - & - & - & 36.1 & 42.7 & 45.1 & 4470\\
GMPHD \cite{GMPHD} & - & - & - & - & 35.6 & 44.7 & 43.5 & 7492\\
\bottomrule
\end{tabularx}
\begin{tablenotes}
\small
\item[*] \textcolor{first}{\textbf{First place}}, \textcolor{second}{\textbf{Second place}}, \textcolor{third}{\textbf{Third place}}
\item[$\dagger$] Uses Tracktor refined detections
\item[$\ddagger$] Uses CenterTrack refined detections
\end{tablenotes}
\label{tab:public_results}
\end{table*}

\begin{table*}[htbp]
\centering
\setlength\tabcolsep{1.5pt}
\caption{MOT Performance comparison on MOT17 Private, MOT20 Private, DanceTrack and BDD100K}
\renewcommand{\arraystretch}{1.2}  
\resizebox{\textwidth}{!}{%
\begin{tabular}{lcccccccccccccccccc}
\toprule
\multirow{2}{*}{Method} & \multicolumn{3}{c}{MOT17 Private} & \multicolumn{3}{c}{MOT20 Private} & \multicolumn{3}{c}{DanceTrack} & \multicolumn{6}{c}{BDD100K} \\
\cmidrule(lr){2-4} \cmidrule(lr){5-7} \cmidrule(lr){8-10} \cmidrule(lr){11-16}
& HOTA $\uparrow$ & MOTA $\uparrow$ & IDF1 $\uparrow$ & HOTA $\uparrow$ & MOTA $\uparrow$ & IDF1 $\uparrow$ & HOTA $\uparrow$ & MOTA $\uparrow$ & IDF1 $\uparrow$ & mHOTA $\uparrow$ & mMOTA $\uparrow$ & mIDF1 $\uparrow$ & HOTA $\uparrow$ & MOTA $\uparrow$ & IDF1 $\uparrow$ \\
\midrule
\textbf{FLASH} & 43.4 & 28.7 & 56.5 & 38.6 & 22.3 & 50.1 & \textcolor{second}{\textbf{62.0}} & 64.1 & \textcolor{first}{\textbf{72.5}} & \textcolor{first}{\textbf{53.7}} & -111.6 & 47.8 & 58.8 & 11.1 & 55.8 \\
SPAM \cite{spam2024eccv} & \textcolor{first}{\textbf{67.5}} & \textcolor{third}{\textbf{80.7}} & \textcolor{first}{\textbf{84.6}} & \textcolor{first}{\textbf{65.8}} & \textcolor{third}{\textbf{76.5}} & \textcolor{first}{\textbf{81.9}} & \textcolor{first}{\textbf{64.0}} & \textcolor{third}{\textbf{89.2}} & 63.4 & - & - & - & - & - & - \\
SUSHI \cite{SUSHI} & \textcolor{second}{\textbf{66.5}} & \textcolor{second}{\textbf{81.1}} & \textcolor{second}{\textbf{83.1}} & \textcolor{second}{\textbf{64.3}} & 74.3 & \textcolor{second}{\textbf{79.8}} & \textcolor{third}{\textbf{63.3}} & 88.7 & 63.4 & - & - & - & - & - & - \\
GHOST \cite{GHOSTTracker} & 62.8 & 78.7 & 77.1 & \textcolor{third}{\textbf{61.2}} & 73.7 & 75.2 & 56.7 & \textcolor{first}{\textbf{91.3}} & 57.7 & \textcolor{third}{\textbf{45.7}} & \textcolor{second}{\textbf{44.9}} & \textcolor{first}{\textbf{55.6}} & \textcolor{first}{\textbf{61.7}} & \textcolor{second}{\textbf{68.1}} & \textcolor{second}{\textbf{70.9}} \\
ByteTrack \cite{ByteTrack} & 62.8 & 78.9 & 77.1 & 60.4 & 74.2 & 74.5 & 47.7 & \textcolor{second}{\textbf{89.6}} & 53.9 & \textcolor{second}{\textbf{45.4}} & \textcolor{first}{\textbf{45.2}} & \textcolor{second}{\textbf{54.6}} & \textcolor{second}{\textbf{61.6}} & \textcolor{first}{\textbf{68.7}} & \textcolor{third}{\textbf{70.2}} \\
MotionTrack \cite{MotionTrack} & \textcolor{third}{\textbf{65.1}} & \textcolor{second}{\textbf{81.1}} & \textcolor{third}{\textbf{80.1}} & 62.8 & \textcolor{second}{\textbf{78.0}} & 76.5 & - & - & - & - & - & - & - & - & - \\
UTM \cite{UTM} & 64.0 & \textcolor{first}{\textbf{81.8}} & 78.7 & 62.5 & \textcolor{first}{\textbf{78.2}} & \textcolor{third}{\textbf{76.9}} & - & - & - & - & - & - & - & - & - \\
QDTrack \cite{QDTrack} & 63.5 & 78.7 & 77.5 & 60.0 & 74.7 & 73.8 & 54.2 & 87.7 & \textcolor{third}{\textbf{50.4}} & 41.7 & 36.3 & 51.5 & \textcolor{third}{\textbf{60.9}} & \textcolor{third}{\textbf{63.7}} & \textcolor{first}{\textbf{71.4}} \\
MOTR \cite{MOTR} & 57.8 & 68.6 & 73.4 & - & - & - & 54.2 & 79.7 & \textcolor{second}{\textbf{51.5}} & - & 32.0 & 43.5 & - & - & - \\
FairMOT \cite{FairMOT} & 59.3 & 73.7 & 72.3 & 54.6 & 61.8 & 67.3 & 39.7 & 82.2 & 40.8 & - & - & - & - & - & - \\
TrackFormer \cite{TrackFormer} & 57.3 & 74.1 & 68.0 & 54.7 & 65.7 & 68.6 & - & - & - & - & - & - & - & - & - \\
MeMOT \cite{MeMOT} & 56.9 & 72.5 & 69.0 & 54.1 & 66.1 & 63.7 & - & - & - & - & - & - & - & - & - \\
TETer \cite{TETer} & - & - & - & - & - & - & - & - & - & - & \textcolor{third}{\textbf{39.1}} & \textcolor{third}{\textbf{53.3}} & - & - & - \\
Yu et al. \cite{BDD100K} & - & - & - & - & - & - & - & - & - & - & 25.9 & 44.5 & - & 56.9 & 66.8 \\
\bottomrule
\end{tabular}%
}
\begin{tablenotes}
\small
\item[*] \textcolor{first}{\textbf{First place}}, \textcolor{second}{\textbf{Second place}}, \textcolor{third}{\textbf{Third place}}
\end{tablenotes}
\label{tab:private_results}
\end{table*}

\subsubsection{\textbf{Comparison of FLASH with SoTA MOT methods}}

\textbf{Performance on Public Detection Benchmarks.}
Our experimental evaluation on MOT17 and MOT20 public detection benchmarks reveals distinct performance characteristics of FLASH compared to state-of-the-art methods. As demonstrated in Table \ref{tab:public_results}, FLASH exhibits suboptimal performance on MOT17 public detections, achieving a HOTA score of 43.60\% and MOTA of 34.3\%, substantially lower than leading methods such as GHOST \cite{GHOSTTracker} (50.7\% HOTA, 61.6\% MOTA) and ArTIST-C \cite{ArTIST} (48.9\% HOTA, 62.3\% MOTA). This performance gap can be attributed to the inherent limitations of public detections, which provide lower-quality bounding boxes that adversely affect SAM2's segmentation initialization.
Despite these overall performance limitations, FLASH demonstrates notable identity preservation capabilities, achieving the third-best identity switch count of 1283 on MOT17, surpassed only by GHOST with Tracktor detections (1144) and TrackPool (1188). This superior IDSW performance indicates SAM2's effective memory-based tracking mechanism, which maintains consistent object identities once successfully initialized. However, the relatively modest IDF1 score of 56.9\% reveals that this identity consistency does not translate to high association quality due to the propagation of false positives from poor initial detections.
On MOT20 public detections, FLASH maintains competitive association performance with an IDF1 score of 52.0\%, outperforming several established methods including SORT \cite{SORT} (45.1\%) and GMPHD \cite{GMPHD} (43.5\%). However, the MOTA performance of 24.2\% remains significantly below optimal, reinforcing the critical dependency on detection quality for overall tracking performance.

\textbf{Performance on Private Detection Benchmarks.}

The evaluation on private detection benchmarks, as presented in Table \ref{tab:private_results}, utilizing high-quality detectors, provides crucial insights into FLASH's fundamental capabilities when detection quality constraints are minimized. On MOT17 private detections, FLASH achieves a HOTA of 43.4\% and MOTA of 28.7\%, substantially trailing leading methods such as SPAM \cite{spam2024eccv} (67.5\% HOTA, 80.7\% MOTA) and SUSHI \cite{SUSHI} (66.5\% HOTA, 81.1\% MOTA). This persistent performance gap, despite improved detection quality, reveals inherent limitations in FLASH's approach to crowded pedestrian tracking scenarios when detections are partially occluded.

On MOT20 private detections, FLASH maintains consistent but suboptimal performance with a HOTA of 38.6\% and MOTA of 22.3\%, significantly lower than top-performing methods including SPAM \cite{spam2024eccv} (65.8\% HOTA, 76.5\% MOTA), SUSHI \cite{SUSHI} (64.3\% HOTA, 74.3\% MOTA), and UTM (62.5\% HOTA, 78.2\% MOTA). The IDF1 score of 50.1\% on MOT20 shows relatively better identity preservation compared to detection accuracy, though still trailing SPAM \cite{spam2024eccv} (81.9\%) and SUSHI \cite{SUSHI} (79.8\%) by substantial margins. This performance pattern across both MOT17 and MOT20 private benchmarks reinforces FLASH's fundamental challenge in dense pedestrian scenarios, where precision in detection and association becomes critical.

Notably, FLASH demonstrates exceptional performance on the DanceTrack dataset, achieving the highest IDF1 score of 72.5\% among all evaluated methods and ranking second in HOTA (62.0\%). This outstanding identity preservation performance validates SAM2's memory-based tracking capabilities in scenarios where objects maintain distinctive visual characteristics. The superior IDF1 performance, combined with competitive HOTA scores, demonstrates FLASH's particular strength in maintaining long-term identity consistency across temporal sequences.

On the BDD100K validation set, FLASH achieves remarkable class-balanced performance with the highest mHOTA score of 53.7\%, indicating superior tracking quality when all object categories are weighted equally. This performance suggests that SAM2's temporal consistency mechanisms are particularly effective across diverse object types, including pedestrians, vehicles, and cyclists.

\subsubsection{\textbf{Understanding FLASH's Variable Performance}}

FLASH's performance varies dramatically across different datasets. While it achieves exceptional results on DanceTrack, its performance is poor on MOT17 and only moderate on MOT20 and BDD100K. Although FLASH is designed to maintain consistent ID tracking, its inconsistent performance across datasets stems from three primary factors:

\begin{enumerate}[leftmargin=*]
    \item \textbf{False Positives from Poor Initial Bounding Boxes}: The low performances across different metrics are primarily a consequence of the high numbers of false positives. These emerge from poor provided initial bounding boxes that, when given to SAM2, inevitably result in trackings of wrong objects rather than target classes. This issue is especially pronounced when bounding boxes are provided for partially or fully occluded objects, leading to artifacts that track part of the front object (rather than the occluded one) or even causing two objects to be assigned to one tracklet. This problem persists particularly when such bounding boxes pass through the online association without being filtered out.

    \item \textbf{Discrepancies in Occlusion Handling}: In MOT17 and MOT20 datasets, ground truth is provided for occluded objects. However, since FLASH converts polygonlets of SAM2 to bounding boxes, it fails to provide any bounding box when objects are fully occluded, or the converted bounding boxes for partially occluded objects are not aligned with the ground truth bounding boxes in MOT challenge. This discrepancy significantly reduces evaluation metrics even when objects are actually being tracked correctly.

    \item \textbf{SAM2's Architectural Limitations with Similar Objects}: When objects with similar appearances interact closely or occlude one another, SAM2's memory-based tracking cannot effectively distinguish between them, leading to identity confusion and tracking failures. This limitation is intrinsic to SAM2's architecture, which relies heavily on appearance and spatial cues that become ambiguous when similar-looking objects interact.
    
\end{enumerate}

\subsection{Sweeping Detections through Time (SAM2Auto)}

\subsubsection{\textbf{Efficacy of the Complete Pipeline}}
The integration of SMART-OD with the FLASH tracker demonstrates the effectiveness of our complete SAM2Auto pipeline, transforming detection-only results into a robust multi-object tracking system through memory-based persistence. This capability makes the pipeline a reliable engine for auto-annotation applications.

Following our ablation study of SMART-OD's detection performance (detailed in Appendix \ref{appendix:experiments_details}), we applied FLASH to the final detected objects from SMART-OD on the MOT17 training set to evaluate the effectiveness of the full SAM2Auto pipeline. Table \ref{SMARTOD_SAM2Auto_comparison} presents the comprehensive performance comparison, highlighting substantial improvements across multiple evaluation metrics.


\begin{enumerate}[leftmargin=*]
    \item \textbf{Memory-Based Tracking: The Key Innovation}: The core advantage of SAM2Auto lies in its memory-based tracking mechanism: \textbf{once an object is detected by SMART-OD for the first time, it will be detected and tracked persistently throughout the sequence}. This eliminates the need for re-detection in subsequent frames and fundamentally transforms tracking from frame-by-frame detection to persistent object maintenance. 
    
    The dramatic improvement in MOTA from -0.014 to 0.181 (1,393\% increase) directly demonstrates this capability. While SMART-OD's high precision (72.8\%) provides quality initial detections, FLASH's SAM2 memory ensures that these objects remain tracked regardless of subsequent visual challenges. The 52.2\% recall improvement reflects how memory-based tracking recovers objects that would be missed in individual frames.
    
    \item \textbf{Strategic Design Validation}: Our precision-focused detection strategy proves optimal for memory-based tracking. The controlled increases in false positives (22.6\%) and false negatives (31.2\%) are substantially offset by the memory system's persistent tracking capability. The DetA improvement of 206\% demonstrates how memory-based continuity enhances overall system performance beyond individual frame detection.
    
    The complete SAM2Auto pipeline achieves its design objective through a two-stage approach: \textbf{SMART-OD provides high-precision initial detection, and FLASH's SAM2 memory ensures persistent tracking thereafter}. This delivers reliable open-vocabulary multi-object tracking without sequence-specific optimization, making it practically applicable for diverse real-world scenarios where continuous object tracking after initial detection is crucial.
\end{enumerate}

\begin{table}[ht]
\centering
\setlength\tabcolsep{2pt}
\renewcommand{\arraystretch}{1.4}  
\caption{Performance comparison between SMART-OD detection-only and complete SAM2Auto pipeline on MOT17 training set}
\label{SMARTOD_SAM2Auto_comparison}
\begin{tabularx}{\columnwidth}{l *{3}{>{\centering\arraybackslash}X}}
\toprule
\textbf{Metric} & \textbf{SMART-OD} & \textbf{SAM2Auto} & \textbf{Improvement} \\
\midrule
MOTA & -0.014 & 0.181 & \cellcolor{green!20}\textbf{+0.195} (1,393\%) \\
Precision & 0.728 & 0.828 & \textbf{+0.100} (13.7\%) \\
Recall & 0.224 & 0.341 & \textbf{+0.117} (52.2\%) \\
False Positives & 2,516 & 3,084$^*$ & \textbf{+568} (22.6\%) \\
False Negatives & 24,206 & 31,770$^*$ & \textbf{+7,564} (31.2\%) \\
DetA & 0.084 & 0.257 & \cellcolor{green!20}{\textbf{+0.173} (206\%)} \\
IDF1 & N/A & 1.156 & \cellcolor{green!20}\textbf{New metric} \\
HOTA & N/A & 0.406 & \cellcolor{green!20}{\textbf{New metric}} \\
\bottomrule
\end{tabularx}
\\
\vspace{4pt}  
\footnotesize{$^*$Average per sequence values from tracking results}
\end{table}

\begin{table}[htbp]
\centering
\caption{Comparison Criteria of Annotation Methods}
\label{comparison_criteria}
\renewcommand{\arraystretch}{1.3}
\begin{tabularx}{\columnwidth}{l*{4}{>{\centering\arraybackslash}X}}
\toprule
\textbf{Method} & \textbf{Label Type} & \textbf{Label Effort} & \textbf{Tracker} & \textbf{Training Required} \\
\midrule
\textbf{SAM2Auto} & SMART-OD (Auto) & \cellcolor{green!20}0\% & SAM2ASH & \cellcolor{green!20}No \\
FLASH & Curated & \cellcolor{yellow!20}0\% & SAM2ASH & \cellcolor{green!20}No \\
\multirow{2}{*}{SPAMming \cite{spam2024eccv}} & \multirow{2}{*}{SPAM} & \cellcolor{red!20}3.3\% & ByteTrack & \cellcolor{red!20}Yes \\
& & \cellcolor{red!20}3.3\% & GHOST & \cellcolor{red!20}Yes \\
\bottomrule
\end{tabularx}
\end{table}

\begin{table*}[htbp]
\centering
\setlength\tabcolsep{1.5pt}
\caption{Auto Annotation Performance comparison on MOT17 Private, MOT20 Private, DanceTrack and BDD100K}
\renewcommand{\arraystretch}{1.2}  
\resizebox{\textwidth}{!}{%
\begin{tabular}{lcccccccccccccccccc}
\toprule
\multirow{2}{*}{Method} & \multicolumn{3}{c}{MOT17 Private} & \multicolumn{3}{c}{MOT20 Private} & \multicolumn{3}{c}{DanceTrack} & \multicolumn{6}{c}{BDD100K} \\
\cmidrule(lr){2-4} \cmidrule(lr){5-7} \cmidrule(lr){8-10} \cmidrule(lr){11-16}
& HOTA $\uparrow$ & MOTA $\uparrow$ & IDF1 $\uparrow$ & HOTA $\uparrow$ & MOTA $\uparrow$ & IDF1 $\uparrow$ & HOTA $\uparrow$ & MOTA $\uparrow$ & IDF1 $\uparrow$ & mHOTA $\uparrow$ & mMOTA $\uparrow$ & mIDF1 $\uparrow$ & HOTA $\uparrow$ & MOTA $\uparrow$ & IDF1 $\uparrow$ \\
\midrule
\textbf{SAM2Auto} & 40.5 & 10.3 & 50.0 & 32.3 & 11.6 & 40.8 & 37.4 & -92.6 & 36.7 & \cellcolor{yellow!50}38.9 & \cellcolor{red!50}-271.6 & \cellcolor{yellow!50}32.6 & \cellcolor{green!50}56.6 & \cellcolor{red!20}8.3 & \cellcolor{green!50}51.8 \\
FLASH & \textcolor{third}{\textbf{43.4}} & \textcolor{third}{\textbf{28.7}} & \textcolor{third}{\textbf{56.5}} & \textcolor{third}{\textbf{38.6}} & \textcolor{third}{\textbf{22.3}} & \textcolor{third}{\textbf{50.1}} & \textcolor{first}{\textbf{62.0}} & \textcolor{third}{\textbf{64.1}} & \textcolor{first}{\textbf{72.5}} & \cellcolor{green!20}53.7 & \cellcolor{red!50}-111.6 & \cellcolor{green!20}47.8 & \cellcolor{green!50}58.8 & \cellcolor{red!20}11.1 & \cellcolor{green!50}55.8 \\
SPAM-GHOST \cite{spam2024eccv} & \textcolor{second}{\textbf{51.3}} & \textcolor{second}{\textbf{61.9}} & \textcolor{second}{\textbf{62.1}} & \textcolor{second}{\textbf{47.0}} & \textcolor{first}{\textbf{58.2}} & \textcolor{second}{\textbf{60.7}} & \textcolor{second}{\textbf{41.0}} & \textcolor{second}{\textbf{76.3}} & \textcolor{third}{\textbf{44.8}} & - & - & - & - & - & - \\
SPAM-ByteTrack \cite{spam2024eccv} & \textcolor{first}{\textbf{51.6}} & \textcolor{first}{\textbf{64.0}} & \textcolor{first}{\textbf{63.0}} & \textcolor{first}{\textbf{47.9}} & \textcolor{second}{\textbf{57.6}} & \textcolor{first}{\textbf{61.4}} & \textcolor{third}{\textbf{39.5}} & \textcolor{first}{\textbf{76.4}} & \textcolor{second}{\textbf{45.0}} & - & - & - & - & - & - \\
\bottomrule
\end{tabular}%
}
\begin{tablenotes}
\small
\item[*] \textcolor{first}{\textbf{First place}}, \textcolor{second}{\textbf{Second place}}, \textcolor{third}{\textbf{Third place}}
\end{tablenotes}
\label{tab:Auto_private_results}
\end{table*}

\subsubsection{\textbf{Differentiating SAM2Auto from SPAMming}}

Since SPAMming \cite{spam2024eccv} is the only closest method to ours in terms of automated annotations, it is important to highlight the fundamental differences between SAM2Auto and SPAMming, as summarized in Table \ref{comparison_criteria}, before comparing the results. These methodological differences directly impact the tracking performance and explain the variations in our experimental results. For fairness of comparison, we only consider their performance with the least annotation effort ( 3.3\%). We also consider the results of FLASH as the special case of SAM2Auto when labels are provided.

\textbf{Annotation Strategy.} The most significant distinction lies in the annotation approach. While SPAMming leverages 3.3\% of manually annotated ground truth data to train and fine-tune both its components (labeling engine and tracker), SAM2Auto operates in a completely annotation-free manner, relying solely on SMART-OD for automatic object detection. This zero-annotation constraint, while eliminating manual labeling costs, inherently limits the method's ability to adapt to dataset-specific characteristics that manual annotations would provide. In case of FLASH, the curated annotation is provided from previous tracking studies where models were trained using 100\% of the ground truth annotated training set, but not the test set that is the subject of evaluation.

\textbf{Training and Adaptation.} SPAMming fine-tunes both its labeling pipeline and tracking components (ByteTrack or GHOST) using the available ground truth annotations, allowing the system to adapt to the target dataset's specific object appearances and motion patterns. In contrast, SAM2Auto uses the pre-trained SMART-OD and SAM2ASH tracker without any dataset-specific training, making it more generalizable but potentially less optimal for specific scenarios.

\textbf{Tracking Architecture.} The tracking components also differ fundamentally in their design philosophy. SPAMming employs conventional online MOT trackers (ByteTrack/GHOST) that maintain object predictions even during occlusions through motion models and appearance features. Additionally, these trackers are fine-tuned with low and high confidence thresholds to retain or remove tracklets based on their confidence scores and tracklet lifespans.
SAM2ASH, however, is a memory-based tracker that generates masklets based on visual appearance and does not provide predictions during full occlusions, which impacts tracking metrics in datasets with ground truth for occluded objects. More precisely, while GHOST is described as an online tracker, it actually employs offline association by adapting its re-identification weights using the full sequence. In contrast, SAM2ASH operates as an offline tracker with online object association, where ByteTrack is used to reduce the computational overhead of overall tracking by avoiding processing redundant objects. Notably, ByteTrack's parameters in SAM2ASH are configured not to filter out any objects based on confidence scores, as we expect to receive high-quality objects from the labeling component.

These architectural and methodological differences establish the context for understanding the performance trade-offs observed in our experimental evaluation, where the fully automated nature of SAM2Auto comes at the cost of some tracking accuracy compared to methods that utilize manual annotations.

\subsubsection{\textbf{SAM2Auto Results}}

The experimental results provide a comprehensive view of the performance trade-offs between tracking capability and detection quality in automated annotation systems. By comparing SAM2Auto, FLASH, and SPAMming, we can isolate the impact of each component and understand the path toward fully automated tracking.

\textbf{FLASH as Upper Bound for SAM2Auto.} FLASH's performance demonstrates the tracking capability of SAM2Auto when provided with high-quality curated labels. On MOT17, FLASH achieves HOTA of 43.4, MOTA of 28.7, and IDF1 of 56.5, showing that even with good labels, there remains a performance gap compared to SPAMming's fine-tuned trackers (HOTA: 51.6, MOTA: 64.0, and IDF1:63\%). This gap can be attributed to SPAMming's dataset-specific training on 3.3\% ground truth data, which allows its trackers to adapt to the specific motion patterns and appearance characteristics of each dataset. The comparison reveals that while SAM2ASH is a capable tracker, dataset-specific adaptation provides SPAMming with approximately 8-10 HOTA points advantage.

\textbf{SAM2Auto: Measuring Detection Impact.} The performance difference between SAM2Auto and FLASH directly quantifies the impact of automatic detection quality. On MOT17, although there is slight gap in HOTA dropping from 43.4 to 40.5 (3 points), the deficiency in MOTA is more dramatic, falling from 28.7 to 10.3 (18.4 points). This indicates that when SMART-OD's automatic detections introduce few false positives, it can exarcerbate the metrics by tracking them through time along with missed detections compared to curated labels. To achieve fully automated tracking matching SPAMming's performance, the detection component would need to improve by approximately 11 HOTA points (from 40.5 to 51.6) on MOT17. Similar patterns emerge on MOT20, where SAM2Auto would need to improve from 32.3 to 47.9 HOTA to match SPAMming.

\textbf{DanceTrack: SAM2ASH's True Potential.} DanceTrack provides compelling evidence of SAM2ASH's capabilities when ground truth protocols align with its design. FLASH achieves exceptional performance (HOTA: 62.0, MOTA: 64.1, IDF1: 72.5), substantially outperforming SPAMming (HOTA: 39.5-41.0). This dramatic reversal suggests that SAM2ASH excels when tracking protocols do not penalize the absence of predictions for fully occluded objects. The MOT Challenge ground truth protocol, which includes annotations for partially and fully occluded objects, creates a mismatch with SAM2ASH's memory-based design that only provide tracks on visible objects. DanceTrack's protocol, focusing on visible dancers, better aligns with SAM2ASH's strengths, revealing its true tracking potential.

\textbf{BDD100K: Multi-class Challenges.} The BDD100K results further illustrate the detection quality gap. While single-class HOTA scores are relatively close (SAM2Auto: 56.6 vs FLASH: 58.8), the multi-class mMOTA values are severely negative for both methods (-271.6 and -111.6 respectively). This suggests that similar to MOT Challenge datasets, the partial occluded ground truth bounding boxes is penalizing SAM@ASH results. Additionally, automatic detection struggles particularly in multi-class scenarios, though FLASH's better mMOTA indicates that curated labels significantly reduce false positives even in complex driving scenes.

\section{Path to Fully Automated Annotation}
\label{S:5}
The results clearly delineate the roadmap for achieving fully automated annotation comparable to semi-supervised methods. While SMART-OD's ablation study \ref{appendix:experiments_details} demonstrated the detection pipeline's effectiveness, comparing SAM2Auto to FLASH reveals that \textbf{detection quality} remains the primary bottleneck rather than tracking capability. This finding provides clear direction for future improvements in automated tracking systems. Critically, even a few false positives become particularly problematic when combined with strong tracking, as they propagate through the sequence and severely degrade overall performance.

More specifically, \textbf{false positive suppression} emerges as the main challenge for detection in auto-annotation contexts. While detection literature traditionally lay emphasis on recall and precision metrics, the problem of distinguishing genuine objects from spurious detections—and effectively filtering these false candidates—remains understudied. This gap is especially critical for tracking applications, where each false positive can spawn an erroneous tracklet that persists throughout the sequence.

The SAM2Auto pipeline also faces an inherent limitation: objects can only be annotated from the moment SMART-OD first detects them, creating \textbf{annotation gaps between object appearance and initial detection}. While improving SMART-OD to detect objects immediately upon scene entry could address this issue, such perfect detection is unrealistic given practical constraints such as occlusions and adverse lighting conditions. 

A more pragmatic solution involves \textbf{bidirectional tracking}: by applying tracking both forward and backward from the initial detection point, we can recover the complete object trajectory. This approach significantly relaxes the detection requirement rather than demanding a frame-perfect initial detection, and the pipeline only needs to detect each object once throughout the entire sequence. This single-detection requirement is far more achievable and would enable comprehensive annotation coverage while maintaining the fully automated nature of the system.

FLASH's performance analysis, particularly its exceptional results on DanceTrack, reveals important insights about SAM2ASH's capabilities and limitations. While the tracker excels in sequences with clear visibility and moderate object interactions, performance degrades significantly when facing \textbf{frequent occlusions and visually similar objects}. We identify three specific challenges and propose corresponding solutions:

\begin{enumerate}
    \item \textbf{Architectural Enhancements for Visual Similarity}: SAM2's current architecture struggles to distinguish between visually similar objects, particularly during occlusions. Incorporating additional discriminative features—such as motion patterns, temporal consistency, or learned object-specific embeddings—could substantially improve identity preservation in crowded scenes.
    
    \item \textbf{Adaptive Segment Merging}: Despite setting $\tau_{merge}$ conservatively low, redundant segments persist in complex scenarios. An adaptive threshold that adjusts based on scene complexity and object density could eliminate redundancies while preserving distinct objects, directly improving MOTA scores by reducing false positives.
    
    \item \textbf{False Positive Tracklet Suppression}: The current pipeline assumes perfect detections from SMART-OD, requiring manual removal of false positive tracklets—an approach that cannot scale. Implementing automatic false positive suppression, potentially through confidence-based filtering or trajectory analysis, is essential for practical deployment.
\end{enumerate}


Overall, these enhancements—bidirectional tracking, visual similarity handling, and false positive suppression—form a clear pathway toward fully automated tracking that can rival semi-supervised methods while maintaining the crucial advantage of zero manual annotation.

\section{Conclusion}
\label{S:6}
In this paper, we presented SAM2Auto, a fully automated video annotation pipeline that eliminates the need for manual labeling in multi-object tracking. By combining SMART-OD for automatic object detection with FLASH for memory-based tracking, we demonstrated the feasibility of zero-annotation tracking across diverse datasets. Crucially, our entire pipeline operates without any training or fine-tuning on the target datasets, relying solely on pre-trained models, yet still achieves competitive performance. We also introduced FLASH, which bridges video object segmentation and multi-object tracking by providing high-quality curated labels, demonstrating the upper-bound performance of our tracking component when paired with accurate detections.

Our comprehensive evaluation isolated the impact of detection quality versus tracking capability, revealing that detection—particularly false positive suppression—remains the primary bottleneck. While SAM2Auto achieves respectable performance, the gap to semi-supervised methods is well-defined and addressable. Notably, FLASH's exceptional performance on DanceTrack demonstrates that our tracking architecture can even surpass fine-tuned methods under appropriate evaluation protocols. These findings underscore that competitive fully automated tracking is not only feasible but may require relatively modest improvements in detection quality, bidirectional tracking, and architectural enhancements for visual similarity.

SAM2Auto represents an important step toward democratizing video annotation, making large-scale tracking accessible to researchers and practitioners without the resources for extensive manual labeling or computational training. As detection models continue to improve and the proposed enhancements are implemented, we envision fully automated training-free tracking becoming not just a cost-effective alternative, but the preferred approach for many real-world applications. The future of video understanding lies in systems that can learn and annotate without human intervention, and this work provides both a practical framework and a clear roadmap toward that goal.

\section{Acknowledgement}
\label{S:7}
We gratefully acknowledge the computational resources provided by the Digital Research Alliance of Canada (formerly Compute Canada) through support from the NSERC Alliance program.

\bibliography{myreference}

\begin{thebibliography}{100}

\bibitem{dosovitskiy2021imageworth16x16words}
A.~Dosovitskiy, L.~Beyer, A.~Kolesnikov, D.~Weissenborn, X.~Zhai, T.~Unterthiner, M.~Dehghani, M.~Minderer, G.~Heigold, S.~Gelly, J.~Uszkoreit, and N.~Houlsby, ``An image is worth 16x16 words: Transformers for image recognition at scale,'' 2021.

\bibitem{brown2020languagemodelsfewshotlearners}
T.~B. Brown, B.~Mann, N.~Ryder, M.~Subbiah, J.~Kaplan, P.~Dhariwal, A.~Neelakantan, P.~Shyam, G.~Sastry, A.~Askell, S.~Agarwal, A.~Herbert-Voss, G.~Krueger, T.~Henighan, R.~Child, A.~Ramesh, D.~M. Ziegler, J.~Wu, C.~Winter, C.~Hesse, M.~Chen, E.~Sigler, M.~Litwin, S.~Gray, B.~Chess, J.~Clark, C.~Berner, S.~McCandlish, A.~Radford, I.~Sutskever, and D.~Amodei, ``Language models are few-shot learners,'' 2020.

\bibitem{radford2021learningtransferablevisualmodels}
A.~Radford, J.~W. Kim, C.~Hallacy, A.~Ramesh, G.~Goh, S.~Agarwal, G.~Sastry, A.~Askell, P.~Mishkin, J.~Clark, G.~Krueger, and I.~Sutskever, ``Learning transferable visual models from natural language supervision,'' 2021.

\bibitem{touvron2023llamaopenefficientfoundation}
H.~Touvron, T.~Lavril, G.~Izacard, X.~Martinet, M.-A. Lachaux, T.~Lacroix, B.~Rozière, N.~Goyal, E.~Hambro, F.~Azhar, A.~Rodriguez, A.~Joulin, E.~Grave, and G.~Lample, ``Llama: Open and efficient foundation language models,'' 2023.

\bibitem{bommasani2022opportunitiesrisksfoundationmodels}
R.~Bommasani {\em et~al.}, ``On the opportunities and risks of foundation models,'' 2022.

\bibitem{openai2024gpt4technicalreport}
J.~Achiam, S.~Adler, S.~Agarwal, {\em et~al.}, ``Gpt-4 technical report,'' 2024.

\bibitem{yu2022cocacontrastivecaptionersimagetext}
J.~Yu, Z.~Wang, V.~Vasudevan, L.~Yeung, M.~Seyedhosseini, and Y.~Wu, ``Coca: Contrastive captioners are image-text foundation models,'' 2022.

\bibitem{vaswani2023attentionneed}
A.~Vaswani, N.~Shazeer, N.~Parmar, J.~Uszkoreit, L.~Jones, A.~N. Gomez, L.~Kaiser, and I.~Polosukhin, ``Attention is all you need,'' 2023.

\bibitem{schuhmann2021laion400mopendatasetclipfiltered}
C.~Schuhmann, R.~Vencu, R.~Beaumont, R.~Kaczmarczyk, C.~Mullis, A.~Katta, T.~Coombes, J.~Jitsev, and A.~Komatsuzaki, ``Laion-400m: Open dataset of clip-filtered 400 million image-text pairs,'' 2021.

\bibitem{caron2021emergingpropertiesselfsupervisedvision}
M.~Caron, H.~Touvron, I.~Misra, H.~Jégou, J.~Mairal, P.~Bojanowski, and A.~Joulin, ``Emerging properties in self-supervised vision transformers,'' 2021.

\bibitem{ramirez2024linkedin}
E.~Ramírez, ``I was surprised by the great response to my recent post on small object detection—thank you for all the dms and shares.'' LinkedIn post, August 2024.
\newblock Accessed: August 29, 2024.

\bibitem{touvron2021trainingdataefficientimagetransformers}
H.~Touvron, M.~Cord, M.~Douze, F.~Massa, A.~Sablayrolles, and H.~Jégou, ``Training data-efficient image transformers \& distillation through attention,'' 2021.

\bibitem{Arash}
A.~Rocky, Q.~J. Wu, and W.~Zhang, ``Review of accident detection methods using dashcam videos for autonomous driving vehicles,'' {\em IEEE Transactions on Intelligent Transportation Systems}, vol.~25, no.~8, pp.~8356--8374, 2024.

\bibitem{Everingham2010}
M.~Everingham, L.~V. Gool, C.~K.~I. Williams, J.~Winn, and A.~Zisserman, ``The pascal visual object classes (voc) challenge,'' {\em International Journal of Computer Vision}, vol.~88, no.~2, pp.~303--338, 2010.

\bibitem{girshick2015fastrcnn}
R.~Girshick, ``Fast r-cnn,'' 2015.

\bibitem{howard2017mobilenetsefficientconvolutionalneural}
A.~G. Howard, M.~Zhu, B.~Chen, D.~Kalenichenko, W.~Wang, T.~Weyand, M.~Andreetto, and H.~Adam, ``Mobilenets: Efficient convolutional neural networks for mobile vision applications,'' 2017.

\bibitem{Geiger}
A.~Geiger, P.~Lenz, C.~Stiller, and R.~Urtasun, ``Vision meets robotics: The kitti dataset,'' {\em The International Journal of Robotics Research}, vol.~32, no.~11, pp.~1231--1237, 2013.

\bibitem{yang2022maskedgenerativedistillation}
Z.~Yang, Z.~Li, M.~Shao, D.~Shi, Z.~Yuan, and C.~Yuan, ``Masked generative distillation,'' 2022.

\bibitem{steiner2022trainvitdataaugmentation}
A.~Steiner, A.~Kolesnikov, X.~Zhai, R.~Wightman, J.~Uszkoreit, and L.~Beyer, ``How to train your vit? data, augmentation, and regularization in vision transformers,'' 2022.

\bibitem{dalessandro2024afreecaannotationfreecounting}
A.~D'Alessandro, A.~Mahdavi-Amiri, and G.~Hamarneh, ``Afreeca: Annotation-free counting for all,'' 2024.

\bibitem{lu2023openvocabularypointcloudobjectdetection}
Y.~Lu, C.~Xu, X.~Wei, X.~Xie, M.~Tomizuka, K.~Keutzer, and S.~Zhang, ``Open-vocabulary point-cloud object detection without 3d annotation,'' 2023.

\bibitem{yang2023emernerfemergentspatialtemporalscene}
J.~Yang, B.~Ivanovic, O.~Litany, X.~Weng, S.~W. Kim, B.~Li, T.~Che, D.~Xu, S.~Fidler, M.~Pavone, and Y.~Wang, ``Emernerf: Emergent spatial-temporal scene decomposition via self-supervision,'' 2023.

\bibitem{jin2024stereo4dlearningthings3d}
L.~Jin, R.~Tucker, Z.~Li, D.~Fouhey, N.~Snavely, and A.~Holynski, ``Stereo4d: Learning how things move in 3d from internet stereo videos,'' 2024.

\bibitem{zohar2023fomo}
O.~Zohar, A.~Lozano, S.~Goel, S.~Yeung, and K.-C. Wang, ``Open world object detection in the era of foundation models,'' in {\em arXiv preprint arXiv:2312.05745}, 2023.

\bibitem{YOLO-World}
T.~Cheng, L.~Song, Y.~Ge, W.~Liu, X.~Wang, and Y.~Shan, ``Yolo-world: Real-time open-vocabulary object detection,'' 2024.

\bibitem{Sapiens}
R.~Khirodkar, T.~Bagautdinov, J.~Martinez, S.~Zhaoen, A.~James, P.~Selednik, S.~Anderson, and S.~Saito, ``Sapiens: Foundation for human vision models,'' 2024.

\bibitem{RTDETR}
Y.~Zhao, W.~Lv, S.~Xu, J.~Wei, G.~Wang, Q.~Dang, Y.~Liu, and J.~Chen, ``Detrs beat yolos on real-time object detection,'' 2024.

\bibitem{DINO-X}
T.~Ren, Y.~Chen, Q.~Jiang, Z.~Zeng, Y.~Xiong, W.~Liu, Z.~Ma, J.~Shen, Y.~Gao, X.~Jiang, X.~Chen, Z.~Song, Y.~Zhang, H.~Huang, H.~Gao, S.~Liu, H.~Zhang, F.~Li, K.~Yu, and L.~Zhang, ``Dino-x: A unified vision model for open-world object detection and understanding,'' 2024.

\bibitem{Prompt-guidedDETR}
H.~Song and J.~Bang, ``Prompt-guided detr with roi-pruned masked attention for open-vocabulary object detection,'' {\em Pattern Recognition}, vol.~155, p.~110648, 2024.

\bibitem{DetCLIP}
L.~Yao, J.~Han, Y.~Wen, X.~Liang, D.~Xu, W.~Zhang, Z.~Li, C.~Xu, and H.~Xu, ``Detclip: Dictionary-enriched visual-concept paralleled pre-training for open-world detection,'' 2022.

\bibitem{AligningBagRegions}
S.~Wu, W.~Zhang, S.~Jin, W.~Liu, and C.~C. Loy, ``Aligning bag of regions for open-vocabulary object detection,'' 2023.

\bibitem{DetCLIPv2}
L.~Yao, J.~Han, X.~Liang, D.~Xu, W.~Zhang, Z.~Li, and H.~Xu, ``Detclipv2: Scalable open-vocabulary object detection pre-training via word-region alignment,'' 2023.

\bibitem{FMGS}
X.~Zuo, P.~Samangouei, Y.~Zhou, Y.~Di, and M.~Li, ``Fmgs: Foundation model embedded 3d gaussian splatting for holistic 3d scene understanding,'' 2024.

\bibitem{GLEE}
J.~Wu, Y.~Jiang, Q.~Liu, Z.~Yuan, X.~Bai, and S.~Bai, ``General object foundation model for images and videos at scale,'' 2023.

\bibitem{Region-Aware}
D.~Kim, A.~Angelova, and W.~Kuo, ``Region-aware pretraining for open-vocabulary object detection with vision transformers,'' 2023.

\bibitem{OpenEMMA}
S.~Xing, C.~Qian, Y.~Wang, H.~Hua, K.~Tian, Y.~Zhou, and Z.~Tu, ``Openemma: Open-source multimodal model for end-to-end autonomous driving,'' 2024.

\bibitem{OWL}
M.~Minderer, A.~Gritsenko, A.~Stone, M.~Neumann, D.~Weissenborn, A.~Dosovitskiy, A.~Mahendran, A.~Arnab, M.~Dehghani, Z.~Shen, X.~Wang, X.~Zhai, T.~Kipf, and N.~Houlsby, ``Simple open-vocabulary object detection with vision transformers,'' 2022.

\bibitem{ExploitingUnlabeleddatavision}
S.~Zhao, Z.~Zhang, S.~Schulter, L.~Zhao, V.~K.~B. G, A.~Stathopoulos, M.~Chandraker, and D.~Metaxas, ``Exploiting unlabeled data with vision and language models for object detection,'' 2022.

\bibitem{GroundingDINO}
S.~Liu, Z.~Zeng, T.~Ren, F.~Li, H.~Zhang, J.~Yang, Q.~Jiang, C.~Li, J.~Yang, H.~Su, J.~Zhu, and L.~Zhang, ``Grounding dino: Marrying dino with grounded pre-training for open-set object detection,'' 2024.

\bibitem{SAM2}
N.~Ravi, V.~Gabeur, Y.-T. Hu, R.~Hu, C.~Ryali, T.~Ma, H.~Khedr, R.~R{\"a}dle, C.~Rolland, L.~Gustafson, E.~Mintun, J.~Pan, K.~V. Alwala, N.~Carion, C.-Y. Wu, R.~Girshick, P.~Doll{\'a}r, and C.~Feichtenhofer, ``Sam 2: Segment anything in images and videos,'' {\em arXiv preprint arXiv:2408.00714}, 2024.

\bibitem{SAHI}
F.~C. Akyon, S.~O. Altinuc, and A.~Temizel, ``Slicing aided hyper inference and fine-tuning for small object detection,'' {\em 2022 IEEE International Conference on Image Processing (ICIP)}, pp.~966--970, 2022.

\bibitem{BDD100K}
F.~Yu, H.~Chen, X.~Wang, W.~Xian, Y.~Chen, F.~Liu, V.~Madhavan, and T.~Darrell, ``Bdd100k: A diverse driving dataset for heterogeneous multitask learning,'' 2020.

\bibitem{nuScenes}
H.~Caesar, V.~Bankiti, A.~H. Lang, S.~Vora, V.~E. Liong, Q.~Xu, A.~Krishnan, Y.~Pan, G.~Baldan, and O.~Beijbom, ``nuscenes: A multimodal dataset for autonomous driving,'' 2020.

\bibitem{MultiSensorAdverseWeather}
H.~Zhang, L.~Xiao, X.~Cao, and H.~Foroosh, ``Multiple adverse weather conditions adaptation for object detection via causal intervention,'' {\em IEEE Transactions on Pattern Analysis and Machine Intelligence}, vol.~46, no.~3, pp.~1742--1756, 2024.

\bibitem{EnrichedFeatures}
S.~W. Zamir, A.~Arora, S.~Khan, M.~Hayat, F.~S. Khan, M.-H. Yang, and L.~Shao, ``Learning enriched features for real image restoration and enhancement,'' 2020.

\bibitem{Retinex}
Z.~Cui, K.~Li, L.~Gu, S.~Su, P.~Gao, Z.~Jiang, Y.~Qiao, and T.~Harada, ``You only need 90k parameters to adapt light: A light weight transformer for image enhancement and exposure correction,'' 2022.

\bibitem{Zero-DCE}
M.~Afifi, K.~G. Derpanis, B.~Ommer, and M.~S. Brown, ``Learning multi-scale photo exposure correction,'' 2021.

\bibitem{CycleGAN}
J.-Y. Zhu, T.~Park, P.~Isola, and A.~A. Efros, ``Unpaired image-to-image translation using cycle-consistent adversarial networks,'' 2020.

\bibitem{TrackFormer}
T.~Meinhardt, A.~Kirillov, L.~Leal-Taixe, and C.~Feichtenhofer, ``Trackformer: Multi-object tracking with transformers,'' in {\em The IEEE Conference on Computer Vision and Pattern Recognition (CVPR)}, June 2022.

\bibitem{BoostTrack}
V.~D. Stanojevic and B.~T. Todorovic, ``Boosttrack: boosting the similarity measure and detection confidence for improved multiple object tracking,'' {\em Machine Vision and Applications}, vol.~35, p.~53, April 2024.

\bibitem{ByteTrack}
Y.~Zhang, P.~Sun, Y.~Jiang, D.~Yu, F.~Weng, Z.~Yuan, P.~Luo, W.~Liu, and X.~Wang, ``Bytetrack: Multi-object tracking by associating every detection box,'' in {\em Computer Vision -- ECCV 2022} (S.~Avidan, G.~Brostow, M.~Ciss{\'e}, G.~M. Farinella, and T.~Hassner, eds.), (Cham), pp.~1--21, Springer Nature Switzerland, 2022.

\bibitem{DiffMOT}
W.~Lv, Y.~Huang, N.~Zhang, R.-S. Lin, M.~Han, and D.~Zeng, ``Diffmot: A real-time diffusion-based multiple object tracker with non-linear prediction,'' in {\em Proceedings of the IEEE/CVF Conference on Computer Vision and Pattern Recognition}, pp.~19321--19330, 2024.

\bibitem{GHOSTTracker}
J.~Seidenschwarz, G.~Brasó, V.~C. Serrano, I.~Elezi, and L.~Leal-Taixé, ``Simple cues lead to a strong multi-object tracker,'' in {\em 2023 IEEE/CVF Conference on Computer Vision and Pattern Recognition (CVPR)}, pp.~13813--13823, 2023.

\bibitem{ROMOT}
W.~Li, B.~Li, J.~Wang, W.~Meng, J.~Zhang, and X.~Zhang, ``Romot: Referring-expression-comprehension open-set multi-object tracking,'' {\em The Visual Computer}, June 2024.

\bibitem{Samba}
M.~Segu, L.~Piccinelli, S.~Li, Y.-H. Yang, B.~Schiele, and L.~V. Gool, ``Samba: Synchronized set-of-sequences modeling for multiple object tracking,'' 2024.

\bibitem{Sparsetrack}
Z.~Liu, X.~Wang, C.~Wang, W.~Liu, and X.~Bai, ``Sparsetrack: Multi-object tracking by performing scene decomposition based on pseudo-depth,'' {\em IEEE Transactions on Circuits and Systems for Video Technology}, 2025.

\bibitem{SUSHI}
O.~Cetintas, G.~Bras\'o, and L.~Leal-Taix\'e, ``Unifying short and long-term tracking with graph hierarchies,'' in {\em Proceedings of the IEEE/CVF Conference on Computer Vision and Pattern Recognition (CVPR)}, pp.~22877--22887, June 2023.

\bibitem{masa}
S.~Li, L.~Ke, M.~Danelljan, L.~Piccinelli, M.~Segu, L.~Van~Gool, and F.~Yu, ``Matching anything by segmenting anything,'' {\em CVPR}, 2024.

\bibitem{JointModeling}
J.~Zhang, Y.~Cui, G.~Wu, and L.~Wang, ``Joint modeling of feature, correspondence, and a compressed memory for video object segmentation,'' 2023.

\bibitem{PuttingObject}
H.~K. Cheng, S.~W. Oh, B.~Price, J.-Y. Lee, and A.~Schwing, ``Putting the object back into video object segmentation,'' 2024.

\bibitem{TAM-VT}
R.~Goyal, W.-C. Fan, M.~Siam, and L.~Sigal, ``Tam-vt: Transformation-aware multi-scale video transformer for segmentation and tracking,'' 2024.

\bibitem{VideoClick}
N.~Homayounfar, J.~Liang, W.-C. Ma, and R.~Urtasun, ``Videoclick: Video object segmentation with a single click,'' 2021.

\bibitem{TrackingAnything}
H.~K. Cheng, S.~W. Oh, B.~Price, A.~Schwing, and J.-Y. Lee, ``Tracking anything with decoupled video segmentation,'' in {\em ICCV}, 2023.

\bibitem{SegmentMeetsPointTracking}
F.~Rajič, L.~Ke, Y.-W. Tai, C.-K. Tang, M.~Danelljan, and F.~Yu, ``Segment anything meets point tracking,'' 2023.

\bibitem{OVIS}
J.~Qi, Y.~Gao, Y.~Hu, X.~Wang, X.~Liu, X.~Bai, S.~Belongie, A.~Yuille, P.~H.~S. Torr, and S.~Bai, ``Occluded video instance segmentation: A benchmark,'' 2022.

\bibitem{TowardOVVIS}
H.~Wang, C.~Yan, S.~Wang, X.~Jiang, X.~Tang, Y.~Hu, W.~Xie, and E.~Gavves, ``Towards open-vocabulary video instance segmentation,'' 2023.

\bibitem{DynOMo}
J.~Seidenschwarz, Q.~Zhou, B.~Duisterhof, D.~Ramanan, and L.~Leal-Taix{\'e}, ``Dynomo: Online point tracking by dynamic online monocular gaussian reconstruction,'' in {\em International Conference on 3D Vision}, 2025.

\bibitem{SpatialTracker}
Y.~Xiao, Q.~Wang, S.~Zhang, N.~Xue, S.~Peng, Y.~Shen, and X.~Zhou, ``Spatialtracker: Tracking any 2d pixels in 3d space,'' in {\em Proceedings of the IEEE/CVF Conference on Computer Vision and Pattern Recognition (CVPR)}, 2024.

\bibitem{Track4Gen}
H.~Jeong, C.-H.~P. Huang, J.~C. Ye, N.~Mitra, and D.~Ceylan, ``Track4gen: Teaching video diffusion models to track points improves video generation,'' 2024.

\bibitem{CoTracker3}
N.~Karaev, I.~Makarov, J.~Wang, N.~Neverova, A.~Vedaldi, and C.~Rupprecht, ``Cotracker3: Simpler and better point tracking by pseudo-labelling real videos,'' in {\em Proc. {arXiv:2410.11831}}, 2024.

\bibitem{OmniTracker}
J.~Wang, Z.~Wu, D.~Chen, C.~Luo, X.~Dai, L.~Yuan, and Y.-G. Jiang, ``Omnitracker: Unifying visual object tracking by tracking-with-detection,'' {\em IEEE Transactions on Pattern Analysis and Machine Intelligence}, pp.~1--15, 2025.

\bibitem{spam2024eccv}
O.~Cetintas, T.~Meinhardt, G.~BrasÃ³, and L.~Leal-TaixÃ©, ``Spamming labels: Efficient annotations for the trackers of tomorrow,'' in {\em European Conference on Computer Vision (ECCV)}, 2024.

\bibitem{BetterCallSAL}
A.~Osep, T.~Meinhardt, F.~Ferroni, N.~Peri, D.~Ramanan, and L.~Leal-TaixÃ©, ``Better call sal: Towards learning to segment anything in lidar,'' in {\em European Conference on Computer Vision (ECCV)}, 2024.

\bibitem{EfficientTrackAnything}
Y.~Xiong, C.~Zhou, X.~Xiang, L.~Wu, {\em et~al.}, ``Efficient track anything,'' {\em preprint arXiv:2411.18933}, 2024.

\bibitem{ReferEverything}
A.~Bagchi, Z.~Bao, Y.-X. Wang, P.~Tokmakov, and M.~Hebert, ``Refereverything: Towards segmenting everything we can speak of in videos,'' 2024.

\bibitem{SAMURAI}
C.-Y. Yang, H.-W. Huang, W.~Chai, Z.~Jiang, and J.-N. Hwang, ``Samurai: Adapting segment anything model for zero-shot visual tracking with motion-aware memory,'' 2024.

\bibitem{SMITE}
A.~Alimohammadi, S.~Nag, S.~A. Taghanaki, A.~Tagliasacchi, G.~Hamarneh, and A.~M. Amiri, ``Smite: Segment me in time,'' 2024.

\bibitem{MILAutoAnnot}
J.~Wu, Y.~Yu, C.~Huang, and K.~Yu, ``Deep multiple instance learning for image classification and auto-annotation,'' in {\em 2015 IEEE Conference on Computer Vision and Pattern Recognition (CVPR)}, pp.~3460--3469, 2015.

\bibitem{LabelingCosts}
I.~Elezi, Z.~Yu, A.~Anandkumar, L.~Leal-Taixe, and J.~M. Alvarez, ``Not all labels are equal: Rationalizing the labeling costs for training object detection,'' 2021.

\bibitem{OVODScaling}
M.~Minderer, A.~Gritsenko, and N.~Houlsby, ``Scaling open-vocabulary object detection,'' 2024.

\bibitem{DetCLIPv3}
L.~Yao, R.~Pi, J.~Han, X.~Liang, H.~Xu, W.~Zhang, Z.~Li, and D.~Xu, ``Detclipv3: Towards versatile generative open-vocabulary object detection,'' 2024.

\bibitem{APOVIS}
Q.~Ma, S.~Yang, L.~Zhang, Q.~Lan, D.~Yang, H.~Chen, and Y.~Tan, ``Apovis: Automated pixel-level open-vocabulary instance segmentation through integration of pre-trained vision-language models and foundational segmentation models,'' {\em Image and Vision Computing}, vol.~154, p.~105384, 2025.

\bibitem{RealTimeVOS}
B.~Yan, M.~Sundermeyer, D.~J. Tan, H.~Lu, and F.~Tombari, ``Towards real-time open-vocabulary video instance segmentation,'' 2024.

\bibitem{AIDE}
M.~Liang, J.-C. Su, S.~Schulter, S.~Garg, S.~Zhao, Y.~Wu, and M.~Chandraker, ``Aide: An automatic data engine for object detection in autonomous driving,'' 2024.

\bibitem{LidarAutoLabel}
A.~J. Yang, S.~C. Romero, M.~Dvornik, S.~Segal, {\em et~al.}, ``Automatic labeling of objects from lidar point clouds via trajectory-level refinement,'' December 2024.

\bibitem{CosmosAI}
N.~Agarwal {\em et~al.}, ``Cosmos world foundation model platform for physical ai,'' 2025.

\bibitem{RoboflowAutoLabel}
J.~Witt, ``Launch: Auto label images with roboflow,'' {\em Roboflow Blog}, March 2024.

\bibitem{SemiSupervisedOWOD}
S.~S. Mullappilly, A.~S. Gehlot, R.~M. Anwer, F.~Shahbaz~Khan, and H.~Cholakkal, ``Semi-supervised open-world object detection,'' {\em Proceedings of the AAAI Conference on Artificial Intelligence}, vol.~38, p.~4305–4314, Mar. 2024.

\bibitem{DBSCAN}
M.~Ester, H.-P. Kriegel, J.~Sander, and X.~Xu, ``A density-based algorithm for discovering clusters in large spatial databases with noise,'' in {\em Proceedings of the Second International Conference on Knowledge Discovery and Data Mining}, KDD'96, p.~226–231, AAAI Press, 1996.

\bibitem{MOT17}
A.~Milan, L.~Leal-Taixe, I.~Reid, S.~Roth, and K.~Schindler, ``Mot16: A benchmark for multi-object tracking,'' 2016.

\bibitem{SimpleReID}
S.~Karthik, A.~Prabhu, and V.~Gandhi, ``Simple unsupervised multi-object tracking,'' {\em CoRR}, vol.~abs/2006.02609, 2020.

\bibitem{ArTIST}
F.~S. Saleh, S.~Aliakbarian, H.~Rezatofighi, M.~Salzmann, and S.~Gould, ``Probabilistic tracklet scoring and inpainting for multiple object tracking,'' {\em CoRR}, vol.~abs/2012.02337, 2020.

\bibitem{CenterTrack}
X.~Zhou, V.~Koltun, and P.~Krähenbühl, ``Tracking objects as points,'' 2020.

\bibitem{MOT20}
P.~Dendorfer, H.~Rezatofighi, A.~Milan, J.~Shi, D.~Cremers, I.~Reid, S.~Roth, K.~Schindler, and L.~Leal-Taixé, ``Mot20: A benchmark for multi object tracking in crowded scenes,'' 2020.

\bibitem{DanceTrack}
P.~Sun, J.~Cao, Y.~Jiang, Z.~Yuan, S.~Bai, K.~Kitani, and P.~Luo, ``Dancetrack: Multi-object tracking in uniform appearance and diverse motion,'' in {\em Proceedings of the IEEE/CVF Conference on Computer Vision and Pattern Recognition (CVPR)}, 2022.

\bibitem{MOTA}
R.~Kasturi, D.~B. Goldgof, P.~Soundararajan, V.~Manohar, J.~S. Garofolo, R.~Bowers, M.~Boonstra, V.~N. Korzhova, and J.~Zhang, ``Framework for performance evaluation of face, text, and vehicle detection and tracking in video: Data, metrics, and protocol,'' {\em {IEEE} Trans. Pattern Anal. Mach. Intell.}, vol.~31, no.~2, pp.~319--336, 2009.

\bibitem{IDF1}
E.~Ristani, F.~Solera, R.~Zou, R.~Cucchiara, and C.~Tomasi, ``Performance measures and a data set for multi-target, multi-camera tracking,'' in {\em Computer Vision -- ECCV 2016 Workshops} (G.~Hua and H.~J{\'e}gou, eds.), (Cham), pp.~17--35, Springer International Publishing, 2016.

\bibitem{HOTA}
J.~Luiten, A.~O\v{s}ep, P.~Dendorfer, P.~Torr, A.~Geiger, L.~Leal-Taixé, and B.~Leibe, ``Hota: A higher order metric for evaluating multi-object tracking,'' {\em International Journal of Computer Vision}, vol.~129, p.~548–578, Oct. 2020.

\bibitem{tracktor}
P.~Bergmann, T.~Meinhardt, and L.~Leal{-}Taix{\'{e}}, ``Tracking without bells and whistles,'' in {\em The IEEE International Conference on Computer Vision (ICCV)}, October 2019.

\bibitem{MPNs}
G.~Brasó and L.~Leal-Taixé, ``Learning a neural solver for multiple object tracking,'' in {\em The IEEE Conference on Computer Vision and Pattern Recognition (CVPR)}, June 2020.

\bibitem{YOLOX}
Z.~Ge, S.~Liu, F.~Wang, Z.~Li, and J.~Sun, ``{YOLOX:} exceeding {YOLO} series in 2021,'' {\em CoRR}, vol.~abs/2107.08430, 2021.

\bibitem{TrackPool}
C.~Kim, L.~Fuxin, M.~Alotaibi, and J.~M. Rehg, ``Discriminative appearance modeling with multi-track pooling for real-time multi-object tracking,'' in {\em 2021 IEEE/CVF Conference on Computer Vision and Pattern Recognition (CVPR)}, pp.~9548--9557, 2021.

\bibitem{UNS}
F.~Bastani, S.~He, and S.~Madden, ``Self-supervised multi-object tracking with cross-input consistency,'' in {\em Advances in Neural Information Processing Systems} (M.~Ranzato, A.~Beygelzimer, Y.~Dauphin, P.~Liang, and J.~W. Vaughan, eds.), vol.~34, pp.~13695--13706, Curran Associates, Inc., 2021.

\bibitem{SORT}
A.~Bewley, Z.~Ge, L.~Ott, F.~Ramos, and B.~Upcroft, ``Simple online and realtime tracking,'' in {\em 2016 IEEE International Conference on Image Processing (ICIP)}, IEEE, Sept. 2016.

\bibitem{GMPHD}
N.~L. Baisa, ``Online multi-object visual tracking using a gm-phd filter with deep appearance learning,'' in {\em 2019 22th International Conference on Information Fusion (FUSION)}, pp.~1--8, 2019.

\bibitem{MotionTrack}
Z.~Qin, S.~Zhou, L.~Wang, J.~Duan, G.~Hua, and W.~Tang, ``Motiontrack: Learning robust short-term and long-term motions for multi-object tracking,'' in {\em Proceedings of the IEEE/CVF Conference on Computer Vision and Pattern Recognition (CVPR)}, pp.~17939--17948, June 2023.

\bibitem{UTM}
S.~You, H.~Yao, B.-k. Bao, and C.~Xu, ``Utm: A unified multiple object tracking model with identity-aware feature enhancement,'' in {\em 2023 IEEE/CVF Conference on Computer Vision and Pattern Recognition (CVPR)}, pp.~21876--21886, 2023.

\bibitem{QDTrack}
J.~Pang, L.~Qiu, X.~Li, H.~Chen, Q.~Li, T.~Darrell, and F.~Yu, ``Quasi-dense similarity learning for multiple object tracking,'' in {\em Proceedings of the IEEE/CVF Conference on Computer Vision and Pattern Recognition (CVPR)}, pp.~164--173, June 2021.

\bibitem{MOTR}
F.~Zeng, B.~Dong, Y.~Zhang, T.~Wang, X.~Zhang, and Y.~Wei, ``Motr: End-to-end multiple-object tracking with transformer,'' in {\em Computer Vision -- ECCV 2022} (S.~Avidan, G.~Brostow, M.~Ciss{\'e}, G.~M. Farinella, and T.~Hassner, eds.), (Cham), pp.~659--675, Springer Nature Switzerland, 2022.

\bibitem{FairMOT}
Y.~Zhang, C.~Wang, X.~Wang, W.~Zeng, and W.~Liu, ``Fairmot: On the fairness of detection and re-identification in multiple object tracking,'' {\em International Journal of Computer Vision}, vol.~129, no.~11, pp.~3069--3087, 2021.

\bibitem{MeMOT}
J.~Cai, M.~Xu, W.~Li, Y.~Xiong, W.~Xia, Z.~Tu, and S.~Soatto, ``Memot: Multi-object tracking with memory,'' in {\em Proceedings of the IEEE/CVF Conference on Computer Vision and Pattern Recognition (CVPR)}, pp.~8090--8100, June 2022.

\bibitem{TETer}
S.~Li, M.~Danelljan, H.~Ding, T.~E. Huang, and F.~Yu, ``Tracking every thing in the wild,'' in {\em Computer Vision -- ECCV 2022} (S.~Avidan, G.~Brostow, M.~Ciss{\'e}, G.~M. Farinella, and T.~Hassner, eds.), (Cham), pp.~498--515, Springer Nature Switzerland, 2022.

\end{thebibliography}
\cleardoublepage
\appendix

\subsection{SMART-OD Mathematical Formulations}
\label{app:mathematical_formulations}

This section provides the complete mathematical formulations for the SMART-OD pipeline described in Section~\ref{sub:detection_pipeline}. The three-stage pipeline consists of segmentation, mask-guided analysis, and robust thresholding, each with specific mathematical definitions and parameter configurations.

\subsubsection{Segmentation Stage}
\label{app:segmentation}

The first stage employs SAM2 for automatic mask generation across all potential objects in the scene. The segmentation process is defined as:

\begin{equation}
    I_{\text{Masked}} = \text{AutoMask}_{\text{SAM2}}(I, \theta_{s}, \theta_{o}, \theta_{n})
    \label{eq:A1}
\end{equation}

\noindent where $I_{\text{Masked}}$ represents the generated instance masks, $\text{AutoMask}_{\text{SAM2}}$ is the SAM2 segmentation function, $I$ is the input image, $\theta_{s}$ is the stability score threshold, $\theta_{o}$ is the stability score offset, and $\theta_{n}$ is the box Non-Maximum Suppression (NMS) threshold parameter.

\subsubsection{Mask-guided Detection Stage}
\label{app:detection}

The second stage performs open-vocabulary object detection using YOLO-World on the masked representations. The detection process is formulated as:

\begin{equation}
    D_{\text{init}} = \text{OVOD}(I_{\text{Masked}}, C, \theta_{c}, \theta_{i}, \theta_{n})
    \label{eq:A2}
\end{equation}

\noindent where $D_{\text{init}}$ represents the initial detections, $\text{OVOD}$ is the YOLO-World open-vocabulary detection function, $I_{\text{Masked}}$ is the mask-annotated image from Eq.~\ref{eq:A1}, $C$ is the set of custom target classes, $\theta_{c}$ is the confidence threshold, $\theta_{i}$ is the IoU threshold, and $\theta_{n}$ is the NMS threshold parameter.

To eliminate unreasonably sized detections, area ratio filtering is applied:

\begin{equation}
    D_{\text{filtered}} = \{d \in D_{\text{init}} \mid \theta_{\text{min}} < \frac{A_d}{A_I} < \theta_{\text{max}}\}
    \label{eq:A3}
\end{equation}

\noindent where $A_d$ is the area of detection $d$, $A_I$ is the total image area, and $\theta_{\text{min}}$ and $\theta_{\text{max}}$ are the minimum and maximum area ratio thresholds, respectively.

Finally, the comprehensive algorithm of SMART-OD for processing every sequence of a dataset is provided in algorithm \ref{alg:sam2ywsahi}:

\begin{algorithm}
\caption{SMART-OD: Object Detection and Verification Pipeline}
\label{alg:sam2ywsahi}
\begin{algorithmic}[1]
\REQUIRE Image $\mathcal{I}$, parameters $\Theta$, thresholds $\delta$
\ENSURE Verified detections $\mathcal{V}_{\text{box}}$, $\mathcal{V}_{\text{class}}$, $\mathcal{V}_{\text{score}}$

\STATE \textbf{// Stage 1: Segmentation}
\STATE $\mathcal{M} \gets \text{AutoMask}_{\text{SAM2}}(\mathcal{I}, \Theta_s, \Theta_o, \Theta_n)$ \hfill $\triangleright$ Generate masks

\STATE \textbf{// Stage 2: Mask-guided Analysis}
\STATE $\mathcal{D}, \mathcal{L} \gets \text{OVOD}(\mathcal{M}, \mathcal{C}, \Theta_c, \Theta_i, \Theta_n)$ \hfill $\triangleright$ Detect objects

\STATE \textbf{// Stage 3: Robust Verification}
\STATE $\mathcal{C} \gets \text{DBSCAN}(\mathcal{D}_{\text{xyxy}}, \epsilon, \mu)$ \hfill $\triangleright$ Cluster detections
\STATE $\mathcal{R} \gets \{[\min_{b \in \mathcal{C}_i} x_1, \min_{b \in \mathcal{C}_i} y_1, \max_{b \in \mathcal{C}_i} x_2, \max_{b \in \mathcal{C}_i} y_2] \mid \mathcal{C}_i \in \mathcal{C}\}$
\STATE $\theta_d \gets \text{DynamicThreshold}(\mathcal{D}_{\text{conf}})$ \hfill $\triangleright$ Adaptive threshold
\STATE $\theta_{\text{final}} \gets \max(\theta_d, \theta_{\text{min}})$
\STATE $\mathcal{V}_{\text{idx}}, \mathcal{V}_{\text{class}}, \mathcal{V}_{\text{box}}, \mathcal{V}_{\text{score}} \gets \emptyset$
\FOR{$r \in \mathcal{R}$}
    \STATE $\mathcal{P} \gets \text{SAHI}(r, \mathcal{M})$ \hfill $\triangleright$ Process ROI
    \FOR{$d \in \mathcal{D}$ where $d \subset r$}
        \IF{$\max_{p \in \mathcal{P}} \text{IoU}(d, p) > \theta_v$ and $c_d > \theta_{\text{final}}$}
            \STATE $\mathcal{V}_{\text{idx}} \gets \mathcal{V}_{\text{idx}} \cup \{d_{\text{idx}}\}$
            \STATE $\mathcal{V}_{\text{class}} \gets \mathcal{V}_{\text{class}} \cup \{d_{\text{class}}\}$
            \STATE $\mathcal{V}_{\text{box}} \gets \mathcal{V}_{\text{box}} \cup \{d_{\text{box}}\}$
            \STATE $\mathcal{V}_{\text{score}} \gets \mathcal{V}_{\text{score}} \cup \{d_{\text{score}}\}$
        \ENDIF
    \ENDFOR
\ENDFOR
\STATE \textbf{return} $\mathcal{V}_{\text{box}}, \mathcal{V}_{\text{class}}, \mathcal{V}_{\text{score}}$
\end{algorithmic}
\end{algorithm}

\subsection{FLASH Implementation Details}
\label{app:Flash_Details}

\subsubsection{Initialization Module}

The module implements a robust three-phase \textbf{checkpoint management} to prevent data loss during save operations:
\begin{enumerate}
    \item Write to temporary file: $f_{temp} \leftarrow \text{serialize}(\mathcal{X})$
    \item Create backup of existing checkpoint: $f_{backup} \leftarrow f_{checkpoint}$ (if exists)
    \item Promote temporary file to checkpoint: $f_{checkpoint} \leftarrow f_{temp}$
\end{enumerate}

\noindent where $\mathcal{X}$ represents the data being checkpointed (the current state of the system), serialize($\mathcal{X}$) converts these data into a format that can be saved on disk. This three-phase approach ensures that there is always at least one valid checkpoint file, even if the system crashes during saving. Checkpoints includes polygon and bounding box video segments, processing progress and tracking information.

In addition, the \textbf{resuming capability} of the module extracts frame numbers from checkpoint filenames to determine the correct starting point according to the saved checkpoint, as follows:

\begin{multline}
\text{StartFrame} = \\
\begin{cases}
-1 & \text{if checkpoint is } ``\text{initial}" \\
\max(v_{frames}) & \text{if checkpoint is } ``\text{final}" \\
N & \text{if checkpoint} \\
  & \text{contains frame number } N
\end{cases}
\end{multline}
\noindent Based on this restoration logic, when loading a checkpoint, the system extracts metadata from its filename to decide where processing should resume: if labeled "initial," processing starts from the beginning (indicated by the special value -1); if labeled "final," processing continues from the last completely processed frame (maximum frame number in $v_{frames})$); otherwise, if the checkpoint contains a specific frame number N in its filename, processing resumes exactly from that frame. This robust mechanism ensures processing continuity across interruptions, which is essential when handling lengthy video sequences that cannot be processed in a single session due to computational or time constraints.

\subsubsection{Online Object Association Module} \par

\paragraph*{\textbf{Temporal Context Processing:}}
The system processes detections differently based on their temporal context. In the first frame, all valid detections are initialized as new objects requiring segmentation. In subsequent frames, detections are first matched against existing tracks using IoU thresholds, and only those without matches are considered new objects. This approach ensures proper initialization and maintains continuity throughout the video sequence.

For all frames, the module begins by extracting valid bounding boxes using robust validation criteria:

\begin{multline}
\text{ValidBox}(B, W, H) = \\
\begin{cases}
\text{true} & \text{if } \text{size}(B) \in [\lambda_{min}, \lambda_{max}]\ \land \\
& \text{boundaries}(B) \subset [m, W-m] \times \\
& [m, H-m]\ \land \\
& \text{aspect}(B) \in [0.2, 5.0] \\
\text{false} & \text{otherwise}
\end{cases}
\end{multline}

\noindent where $\lambda_{min}$ and $\lambda_{max}$ define the allowed size range, $W$ and $H$ represent frame dimensions, $m$ is the margin parameter, and $\text{aspect}(B)$ calculates the box aspect ratio.

\paragraph*{\textbf{Detection-Track Association:}}
The core functionality of this module is establishing temporal consistency through frame-to-frame detection-to-track mapping. This process maintains object identities across the video sequence by systematically associating new detections with existing tracks. 

Formally, the system maintains a mapping $\mathcal{M}: D_j \rightarrow \mathcal{T}$ that assigns each detection index to a track ID:

\begin{multline}
\mathcal{M}(i) =
\begin{cases} 
t_k & \text{if } \exists k : \text{IoU}(B_{\mathcal{T}_k}, B_{d_i}) > \tau \\
    & \text{and } k = \arg\max_{l} \text{IoU}(B_{\mathcal{T}_l}, B_{d_i}) \\
\end{cases}
\end{multline}

\noindent For each detection $d_i \in D_j$ in frame $j$, the system identifies the track with the highest IoU overlap:

\begin{equation}
\text{matched\_track\_idx} = \arg\max_{k} \text{IoU}(B_{\mathcal{T}_k}, B_{d_i})
\end{equation}

\noindent where $B_{\mathcal{T}_k}$ represents the bounding box of track $\mathcal{T}_k$ and $B_{d_i}$ is the bounding box of detection $d_i$. A match is confirmed when the maximum IoU exceeds threshold $\tau$ (typically 0.5):

\begin{equation}
\text{IoU}(B_{\mathcal{T}_{\text{matched\_track\_idx}}}, B_{d_i}) > \tau
\end{equation}

\noindent Upon confirmation, detection $d_i$ is associated with track $\mathcal{T}_{\text{matched\_track\_idx}}$, and the corresponding track ID is assigned to the detection.

\subsubsection{Annotation and Segmentation Handler (ASH)}

\paragraph*{\textbf{Memory-Efficient Processing:}}
For effective real-time performance, ASH implements two key memory optimization strategies:

\textit{Subset Frame Processing:} For a video sequence with frames $\mathcal{F} = \{F_1, F_2, ..., F_T\}$, ASH employs a subset of frames $\mathcal{S}$ at frame $t$ defined as:
$\mathcal{S}_t = \{F_t, F_{t+1}, ..., F_{T}\}$
where $T$ is the total number of frames in the sequence and $t$ is the frame where new objects are initiated by the Online Object Association module. This allows the system to propagate masks from the frame where new objects are detected to the end of the sequence, avoiding excessive memory requirements.

\textit{Batch Object Processing:} To optimize computational efficiency, ASH processes detected objects in batches. The set of new objects $\mathcal{N}_t = \{o_1^t, o_2^t, ..., o_n^t\}$ detected at frame $t$ is divided into batches:

\begin{equation}
\mathcal{B}_k = \{o_{(k-1)\cdot\beta + 1}^t, o_{(k-1)\cdot\beta + 2}^t, ..., o_{\min(k\cdot\beta, n)}^t\}
\end{equation}

\noindent where $n$ is the number of new objects at time $t$, $\beta$ is the batch size parameter, and $k \in \{1, 2, ..., \lceil n/\beta \rceil\}$.
This approach strikes a balance between throughput and memory constraints, particularly important when handling frames with numerous objects on limited GPU hardware. The batch size parameter ($\beta$) controls the number of objects processed simultaneously, balancing memory requirements and computational efficiency.

Thus, for each batch of objects, the memory-based segmentation model is initialized with the appropriate subsequent frames. This state management strategy ensures that each batch starts with a clean model state while maintaining an appropriate video context.

The overall process of FLASH and FLASH in the Chunk-Based Processing for Long Video Sequences are entailed in algorithm \ref{alg:flash_vos} and algorithm \ref{alg:flash_chunk}.

\subsection{SAM2Auto Implementation Details}
\label{appendix:sam2auto_details}

\subsubsection{\textbf{Representative Sequence Selection}}
The first step involves identifying a sequence with the highest object density in the dataset. This "worst-case" sequence provides a challenging test case for parameter optimization. We select the sequence with the maximum number of objects across all sequences in the dataset:

\begin{equation}
S_{rep} = \arg\max_{S \in \mathcal{D}} \max_{f \in S} |O_f|
\end{equation}

where $S$ represents a sequence in dataset $\mathcal{D}$, $|O_f|$ is the number of objects in frame $f$, and $\max_{f \in S} |O_f|$ identifies the frame with the highest object count within each sequence. This selection criterion ensures we choose the sequence containing the single most crowded frame in the entire dataset.

\begin{algorithm}
\caption{FLASH: Annotation and Segmentation Handler for Video Object Segmentation}
\label{alg:flash_vos}
\begin{algorithmic}[1]
\REQUIRE New objects $\mathcal{N}_t$ at frame $t$, video sequence $\mathcal{F} = \{F_1, F_2, ..., F_T\}$, tracking data, segmentation model $\phi$, batch size $\bm{\beta}$, detection-track IoU threshold $\bm{\tau_\text{track-det}}$, redundant segment merge threshold $\bm{\tau_\text{merge}}$, temporal smoothing factor $\bm{\alpha}$, mask content threshold $\bm{\epsilon}$
\ENSURE Video segments (polygons) $\mathcal{V}$
\STATE $\mathcal{V} \gets \emptyset$ \hfill $\triangleright$ Initialize output segments
\STATE $\mathcal{S}_t \gets \{F_t, F_{t+1}, ..., F_T\}$ \hfill $\triangleright$ Process only from current frame to end

\FOR{each batch $\mathcal{B}_k$ of up to $\beta$ objects from $\mathcal{N}_t$}
    \STATE \textbf{// Object ID Assignment via Tracking}
    \FOR{each $o_i^t \in \mathcal{B}_k$}
        \IF{$\exists j : \text{IoU}(b_i^t, b_j^{t-1}) > \tau_\text{track}$}
            \STATE $\text{ID}(o_i^t) \gets \text{track}_j$ \hfill $\triangleright$ Assign existing track ID
        \ELSE
            \STATE $\text{ID}(o_i^t) \gets \max(\text{IDs}) + 1$ \hfill $\triangleright$ Assign new track ID
        \ENDIF
    \ENDFOR
    
    \STATE \textbf{// Mask Propagation}
    \FOR{each $o_i^t \in \mathcal{B}_k$}
        \FOR{$\delta \gets 0$ \TO $(T-t)$}
            \STATE $M_i^{t+\delta} \gets \phi(o_i^t, F_t, F_{t+\delta})$ \hfill $\triangleright$ Generate masks
            \STATE $P_i^{t+\delta} \gets \psi(M_i^{t+\delta})$ \hfill $\triangleright$ Convert to polygons
            \STATE $\mathcal{V} \gets \mathcal{V} \cup \{(t+\delta, \text{ID}(o_i^t), P_i^{t+\delta})\}$
        \ENDFOR
    \ENDFOR
\ENDFOR

\STATE \textbf{// Post-Processing}
\STATE \textbf{1. Remove Empty Masks}
\FOR{each object $o_i$}
    \STATE $\tau(o_i) \gets \max\{t \in [1,T] : \sum_{x,y} M_i^t(x,y) > \epsilon\}$ 
    \STATE Remove $(t, \text{ID}(o_i), P_i^t)$ from $\mathcal{V}$ for all $t > \tau(o_i)$
\ENDFOR

\STATE \textbf{2. Apply Temporal Smoothing}
\FOR{each object $o_i$}
    \FOR{$t \gets 2$ \TO $\tau(o_i)$}
        \STATE $P_i^t \gets \alpha \cdot P_i^t + (1-\alpha) \cdot P_i^{t-1}$
        \STATE Update $(t, \text{ID}(o_i), P_i^t)$ to $(t, \text{ID}(o_i), P_i^t)$ in $\mathcal{V}$
    \ENDFOR
\ENDFOR

\STATE \textbf{3. Merge Redundant Segments}
\FOR{each frame $t \in [1,T]$}
    \FOR{each pair $(P_i^t, P_j^t)$ where $i \neq j$ in frame $t$}
        \IF{$\text{IoU}(P_i^t, P_j^t) > \tau_\text{merge}$}
            \STATE Merge $(t, \text{ID}(o_i), P_i^t)$ and $(t, \text{ID}(o_j), P_j^t)$ in $\mathcal{V}$
        \ENDIF
    \ENDFOR
\ENDFOR

\STATE \textbf{return} $\mathcal{V}$
\end{algorithmic}
\end{algorithm}

\begin{algorithm}
\caption{FLASH: Robust Chunk-Based Processing for Long Video Sequences}
\label{alg:flash_chunk}
\begin{algorithmic}[1]
\REQUIRE Video sequence $\mathcal{F} = \{F_1, F_2, ..., F_T\}$, annotations $\mathcal{A}$, chunk size $\bm{\chi}$, overlap size $\bm{\omega}$
\ENSURE Complete video segments (polygons) $\mathcal{V}$

\STATE $\mathcal{V} \gets \emptyset$ \hfill $\triangleright$ Initialize output segments
\STATE $\mathcal{B} \gets \emptyset$ \hfill $\triangleright$ Initialize bbox segments

\STATE \textbf{// First attempt full processing}
\STATE success $\gets$ TryFullProcessing($\mathcal{F}$, $\mathcal{A}$, $\mathcal{V}$, $\mathcal{B}$)

\IF{not success}
    \STATE \textbf{// Fall back to chunk-based processing}
    \STATE $n \gets \lceil T / (\chi - \omega) \rceil$ \hfill $\triangleright$ Number of chunks
    \STATE current\_frame $\gets 1$
    
    \FOR{$i \gets 1$ \TO $n$}
        \STATE \textbf{// Find optimal chunk boundaries}
        \STATE optimal\_frame $\gets$ FindOptimalFrame(current\_frame, $\omega$, $\mathcal{B}$)
        \STATE chunk\_start $\gets \max(1, \text{optimal\_frame} - \omega)$
        \STATE chunk\_end $\gets \min(T, \text{chunk\_start} + \chi - 1)$
        
        \STATE \textbf{// Determine overlap region}
        \IF{$i > 1$}
            \STATE overlap\_frames $\gets [$ chunk\_start, chunk\_start $+\omega]$
        \ELSE
            \STATE overlap\_frames $\gets \emptyset$
        \ENDIF
        
        \STATE \textbf{// Process current chunk}
        \STATE $\mathcal{V}_i, \mathcal{B}_i \gets$ ProcessChunk($\mathcal{F}$, $\mathcal{A}$, chunk\_start, chunk\_end, overlap\_frames)
        
        \STATE \textbf{// Merge with previous results at overlap}
        \IF{$i > 1$}
            \STATE $\mathcal{V} \gets$ MergeOverlappingSegments($\mathcal{V}$, $\mathcal{V}_i$, overlap\_frames)
            \STATE $\mathcal{B} \gets$ MergeOverlappingBBoxes($\mathcal{B}$, $\mathcal{B}_i$, overlap\_frames)
        \ELSE
            \STATE $\mathcal{V} \gets \mathcal{V}_i$
            \STATE $\mathcal{B} \gets \mathcal{B}_i$
        \ENDIF
        
        \STATE current\_frame $\gets$ chunk\_end $+ 1$
    \ENDFOR
    
    \STATE \textbf{// Apply final post-processing to full sequence}
    \STATE $\mathcal{V} \gets$ PostProcessPolygonSegments($\mathcal{V}$)
    \STATE $\mathcal{V} \gets$ MergeRedundantSegments($\mathcal{V}$)
\ENDIF

\STATE \textbf{return} $\mathcal{V}$
\end{algorithmic}
\end{algorithm}






\subsubsection{\textbf{Parameter Optimization with SMART-OD}}
We apply SMART-OD to the most crowded frame in the selected sequence. The optimization process seeks to maximize the detection of ground truth objects while minimizing false positives. Formally, we optimize the parameter set $\Theta = \{\theta_s, \theta_o, \theta_n, \theta_c, \theta_i, \theta_v\}$ to maximize the objective function:

\begin{equation}
J(\Theta) = \alpha \cdot \text{Recall}(\Theta) + (1-\alpha) \cdot \text{Precision}(\Theta)
\end{equation}

\noindent where $\alpha$ is a weighting factor that balances recall and precision according to application requirements.

\subsubsection{\textbf{Sequence-Level Verification}}
The optimized parameters are then applied to the entire representative sequence. Performance is evaluated using standard metrics:

\begin{equation}
\text{Precision} = \frac{TP}{TP + FP}, \quad \text{Recall} = \frac{TP}{TP + FN}
\end{equation}

\noindent where TP, FP, and FN represent true positives, false positives, and false negatives, respectively. Parameters are refined to ensure both metrics exceed application-specific thresholds.

\subsubsection{\textbf{Cross-Sequence Validation}}
To verify generalization, we apply the configuration to a randomly selected sequence $S_{val}$ from a different part of the dataset. The configuration is considered valid if:

\begin{align}
\min(\text{Precision}(S_{\text{val}}), \text{Recall}(S_{\text{val}})) 
\geq \, \\ \gamma \cdot {} \nonumber
\min(\text{Precision}(S_{\text{rep}}), \text{Recall}(S_{\text{rep}}))
\end{align}

\noindent where $\gamma \leq 1$ is a tolerance factor (typically 0.9).

\subsubsection{\textbf{Dataset-Wide Detection}}
The validated SMART-OD configuration is deployed across the entire dataset using distributed processing. For each sequence $S_i$, the pipeline generates:

\begin{equation}
D_i = \text{SMART-OD}(S_i, \Theta_{opt})
\end{equation}

\noindent where $D_i$ represents the detection results for sequence $S_i$, and $\Theta_{opt}$ is the optimized parameter set.





\subsubsection{\textbf{Full-Scale Sequence Manager Application}}
The Sequence Manager, with FLASH as its core processing component, is applied to the detection results:

\begin{equation}
V_i = \text{SequenceManager}(S_i, D_i, \Theta_{SM})
\end{equation}

\noindent where $V_i$ represents the video segments (polygons) for sequence $S_i$, and $\Theta_{SM}$ is the set of Sequence Manager parameters that control FLASH's behavior and chunking strategy. This integrated approach automatically handles memory-efficient processing across all sequences without requiring manual intervention.

\subsubsection{\textbf{Quality Assurance}}
Final annotations undergo verification on a stratified sample of sequences. For each sampled sequence, we compute the Intersection over Union (IoU) between auto-generated annotations and a small set of manual annotations:

\begin{equation}
\text{IoU}(V_i, M_i) = \frac{|V_i \cap M_i|}{|V_i \cup M_i|}
\end{equation}

\noindent Sequences with IoU below a threshold $\tau_{QA}$ undergo targeted parameter refinement.

\subsection{Experiments' Details}
\label{appendix:experiments_details}

\subsubsection{\textbf{Experimental Setup}} \par

In \textit{SMART-OD}, we carefully configured SAM2's automatic mask generator with optimized parameters to balance recall and precision: stability score threshold of 0.90, stability score offset of 0.7, and box NMS threshold of 0.7. These parameters were determined through extensive experimentation to ensure robust mask generation across diverse scenes. For object detection, we configured YOLO-World to identify eight specific classes in autonomous driving scenarios for BDD (Pedestrian, Rider, Car, Truck, Bus, Train, Motorcycle, Bicycle) and exclusively the person class for MOT17, MOT20, and DanceTrack datasets. 
To maintain consistent performance across all datasets, we employed the following unified parameters: YOLO confidence threshold of 0.001, YOLO IoU threshold of 0.1, YOLO NMS threshold of 0.1, verification IoU threshold of 0.03, minimum area ratio of 0.0008, and maximum area ratio of 0.20. Our dynamic threshold methods include kmeans, mean of standard deviation, kmeans of standard deviations (default), and double kmeans. In the Detection Clustering stage (3.1) of SMART-OD, we set the maximum distance between points ($\epsilon$) to 100 and the minimum number of samples in a cluster ($\mu$) to 1. The pipeline generates standardized MOT format outputs.

For \textit{SAM2ASH}, our parameter optimization focused on balancing segmentation quality with processing efficiency. Key parameters include: batch size (${\beta = 5}$) controlling the number of objects processed simultaneously; chunk size ($\\chi = 50$) frames with overlap size ($\\omega = 10$) frames for efficient sequence processing; 
IoU tracking threshold ($\tau_\text{track-det} = 0.5$) determining when new detections associate with existing tracks during online tracking, while using a separate overlap threshold ($\tau_\text{overlap} = 0.7$) for merging segments across overlapping frames between processing chunks;

and box validation constraints following the criteria established in the method section:
size bounds of $\lambda_\text{min} = 10$ pixels and $\lambda_\text{max} = 1000$ pixels, and a boundary margin $m = 0.5$ pixels from frame edges, and aspect ratio constraints between 0.2 and 5.0, ensuring only well-formed bounding boxes are processed.
We configured ByteTracker with track threshold ($\tau_\text{conf} = 0.6$), match threshold ($\tau_\text{match} = 0.7$), and track buffer of 20 frames for robust ID maintenance. For temporal consistency, we applied adaptive smoothing with base factor ($\alpha = 0.2$) that adjusts dynamically based on object movement speed, reducing jitter while preserving natural motion.
we employed an IoU threshold of $\tau_\text{merge} = 0.3$ to identify and combine overlapping object instances, with minimal mask content threshold $\epsilon = 3$ pixels for determining valid mask representation, used in the empty mask removal process.
When evaluating FLASH as a standalone tracker with refined public and private detections, we set the confidence score of all bounding boxes to the fixed value of 0.95 to make sure that all detections are passed through the ByteTracker's online association and none of them are filtered out. Having said that, when employing FLASH as an integrated component of SAM2Auto, since the confidence scores of detected objects from YOLO-World are usually very low, we map their confidence scores to the range (0.7-0.95) using linear rescaling leading to same condition to that of tracking mode of FLASH:
\begin {equation}
\text{conf}_\text{new} = 0.7 + (\text{conf}_\text{orig} - \text{min}_\text{conf}) \cdot \frac{0.25}{\text{max}_\text{conf} - \text{min}_\text{conf}}
\end{equation}
This ensures consistent tracking conditions across different detection sources and implementation scenarios, while allowing ByteTracker's online association to maintain appropriate confidence-based tracking decisions.

\begin{figure*}[!t]
\includegraphics[trim={0.2cm, 0.2cm, 0.2cm, 0.1cm}, clip, width=18cm]{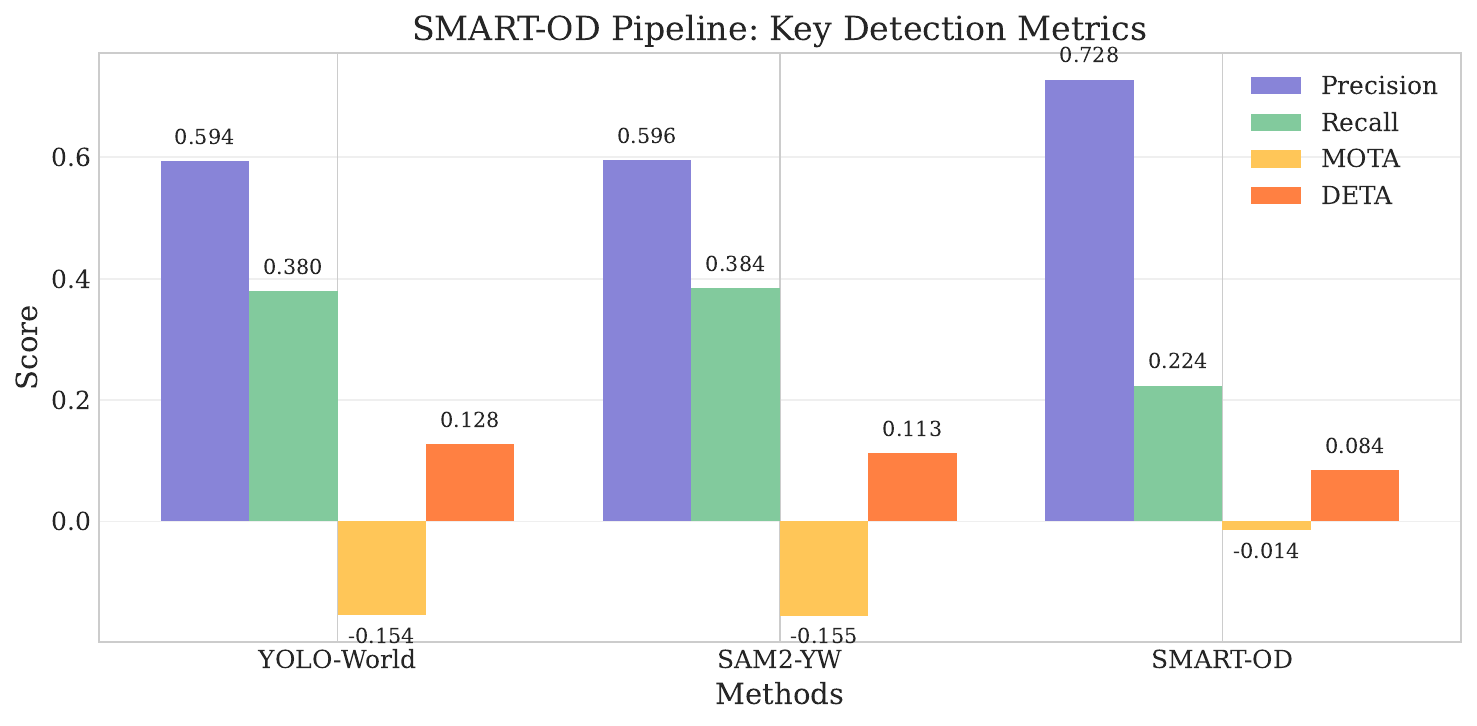}
\caption{Key detection metrics for YOLO-World, SAM2-YW, and SMART-OD on MOT17 training set. SMART-OD achieves the highest precision (0.728) and MOTA (-0.014) by optimizing the precision-recall trade-off.}
\vspace{-0.3cm}
\label{fig5}
\end{figure*}

\subsubsection{\textbf{SMART-OD as Efficient Object Detection}}
\label{SMARTOD_Ablation}
As discussed in \ref{sub:detection_pipeline}, the SMART-OD pipeline aims to provide the most true positive objects in the entire dataset without the need to fine-tune the parameters for each frame or sequence of dataset. For this purpose, we evaluated the impact of each component through an ablation study on MOT17 \cite{MOT17} training sets. As shown in Figure \ref{fig5} and Table \ref{SMART-OD_Results}, we compare the performance of (i) YOLO-World baseline, (ii) SAM2 combined with YOLO-World (SAM2-YW), and (iii) the full SMART-OD pipeline with SAHI verification.

It is important to note that at this detection stage, objects are not assigned persistent identities across frames, which results in lower values for tracking-oriented metrics such as MOTA. The negative MOTA values (-0.154 to -0.014) reflect this limitation, as the metric heavily penalizes identity switches and tracking inconsistencies. However, our primary objective is to evaluate detection performance, making precision and recall the more relevant metrics.

The results demonstrate that the full pipeline successfully achieves its primary objective of maximizing true positive detections while minimizing false positives. The most significant improvement comes from the dramatic reduction in false positives - from 5,861 in the YOLO-World baseline to 2,516 in the full pipeline, representing a 57\% decrease (Table \ref{SMART-OD_Results}). This reduction validates our approach of using a robust verification stage to filter out spurious detections without requiring sequence-specific parameter tuning.
As illustrated in Figure \ref{fig5}, the pipeline exhibits a strategic trade-off between precision and recall. While recall decreases from 38.0\% to 22.4\%, precision substantially improves from 59.4\% to 72.8\%. This trade-off is deliberately engineered into SMART-OD, as the system prioritizes the quality of detections over quantity. The resulting improvement in Multiple Object Tracking Accuracy (MOTA) from -0.154 to -0.014 further confirms that this trade-off leads to better overall detection performance, as MOTA penalizes both false positives and false negatives.

The Detection Accuracy (DETA) metric shows a decrease from 0.128 (YOLO-World) to 0.084 (SMART-OD). This reduction in DETA is expected given the pipeline's design, which prioritizes precision over recall. While a lower DETA might seem counterintuitive, it accurately reflects our deliberate strategy of accepting fewer detections in exchange for greater confidence in those that are made. The substantial improvement in MOTA, combined with the dramatic reduction in false positives, confirms that this trade-off results in superior detection quality for our intended application.

Notably, the intermediate configuration with SAM2 segmentation alone (SAM2-YW) shows minimal improvement over the baseline, with negligible gains in precision and recall (Figure \ref{fig5}) while actually increasing false positives to 6,564 (Table \ref{SMART-OD_Results}). This finding highlights that segmentation alone is insufficient; the true value of SAM2 emerges only when combined with the SAHI verification stage. The effectiveness of the full pipeline stems from the synergistic interaction between the mask-guided analysis provided by SAM2 and the robust thresholding mechanism in SAHI, which adapts to each sequence without manual intervention.
These results empirically demonstrate that SMART-OD successfully provides a dataset-optimized solution for open-vocabulary detection. By accepting a controlled reduction in recall while substantially improving precision (Figure \ref{fig5}) and achieving a 57\% reduction in false positives (Table \ref{SMART-OD_Results}), the pipeline creates a more reliable detection system that generalizes across diverse sequences without per-sequence parameter optimization, thus fulfilling its design goal of efficient, dataset-level object detection.

\begin{table}[ht]
\renewcommand{\arraystretch}{1.5}  
\centering
\caption{False Positive and False Negative counts for SMART-OD components on MOT17 training set}
\label{SMART-OD_Results}
\begin{tabularx}{\columnwidth}{l *{2}{>{\centering\arraybackslash}X}}
\toprule
\textbf{Method} & \textbf{False Positives} & \textbf{False Negatives} \\
\midrule
YOLO-World & 5,861 & 19,411 \\
SAM2-YW & 6,564 & 19,074 \\
\textbf{SMART-OD} & \textbf{2,516} & 24,206 \\
\bottomrule
\end{tabularx}
\end{table}



\end{document}